\documentclass[preprint,10pt,authoryear]{elsarticle}
\usepackage{lineno}
\usepackage{bm}
\usepackage[ruled,linesnumbered]{algorithm2e}
\usepackage{booktabs} 
\usepackage{siunitx} 
\usepackage[table]{xcolor}
\usepackage{color}
\usepackage[left=2.5cm,right=2.5cm,top=3cm,bottom=3cm]{geometry}
\usepackage{svg}
\usepackage{amsmath}

\usepackage{multirow}

\usepackage{booktabs}
\usepackage{graphicx}
\usepackage{subfigure}
\usepackage{epstopdf}
\usepackage{picinpar}
\usepackage{amsmath}
\usepackage{amsfonts}
\usepackage[mathscr]{euscript}
\usepackage{calligra}
\usepackage{array}
\usepackage[authoryear]{natbib} 
\usepackage{setspace}
\usepackage{setspace}
\usepackage{lipsum} 
\usepackage{adjustbox}

\makeatletter
\def\ps@pprintTitle{} 
\makeatother

\usepackage{titlesec}
\titlespacing{\section}{0pt}{6pt}{6pt}
\titlespacing{\subsection}{0pt}{3pt}{3pt}
\titlespacing{\subsubsection}{0pt}{2pt}{2pt}

\DeclareMathAlphabet{\mathpzc}{OT1}{pzc}{m}{it}

\newdefinition{assumption}{Assumption}

\newdefinition{definition}{Definition}
\newdefinition{remark}{Remark}
\newproof{proof}{Proof}

\biboptions{authoryear,compress}

\let\OLDthebibliography\thebibliography
\renewcommand\thebibliography[1]{
  \OLDthebibliography{#1}
  \setlength{\parskip}{0pt}
  \setlength{\itemsep}{0pt plus 0.3ex}
}
\usepackage[colorlinks=true,
            linkcolor=blue,
            citecolor=blue,
            urlcolor=blue]{hyperref}  

\usepackage{cleveref}

\begin{document}

\begin{frontmatter}

\title{SMART: Scalable Multi-Agent Reasoning and Trajectory Planning \newline
in Dense Environments}

\author[a]{Heye Huang}
\ead{hhuang468@wisc.edu}

\author[b]{Yibin Yang\corref{cor1}}
\ead{yyb19@mails.tsinghua.edu}

\author[c]{Wang Chen}
\ead{wchen22@connect.hku.hk}

\author[d]{Tiantian Chen}
\ead{nicole.chen@kaist.ac.kr}

\author[a]{Xiaopeng Li}
\ead{xli2485@wisc.edu}

\author[a]{Sikai Chen\corref{cor1}}
\ead{sikai.chen@wisc.edu}

\address[a]{Department of Civil and Environmental Engineering University of Wisconsin-Madison Madison, Madison, WI, 53706, USA}
\address[b]{School of Vehicle and Mobility, Tsinghua University, Beijing, 100084, China}
\address[c]{Department of Civil Engineering, The University of Hong Kong, Hong Kong, China}
\address[d]{Cho Chun Shik Graduate School of Mobility, Korea Advanced Institute of Science and Technology, South Korea}

\cortext[cor1]{Corresponding author}

%

\begin{abstract}
Multi-vehicle trajectory planning is a non-convex problem that becomes increasingly difficult in dense environments due to the rapid growth of collision constraints. Efficient exploration of feasible behaviors and resolution of tight interactions are essential for real-time, large-scale coordination. This paper introduces SMART, Scalable Multi-Agent Reasoning and Trajectory Planning, a hierarchical framework that combines priority-based search with distributed optimization to achieve efficient and feasible multi-vehicle planning. The upper layer explores diverse interaction modes using reinforcement learning-based priority estimation and large-step hybrid A* search, while the lower layer refines solutions via parallelizable convex optimization. By partitioning space among neighboring vehicles and constructing robust feasible corridors, the method decouples the joint non-convex problem into convex subproblems solved efficiently in parallel. This design alleviates the step-size trade-off while ensuring kinematic feasibility and collision avoidance. Experiments show that SMART consistently outperforms baselines. On $50$\,m $\times$ $50$\,m maps, it sustains over $90\%$ success within $1$\,s up to $25$ vehicles, while baselines often drop below $50\%$. On $100$\,m $\times$ $100$\,m maps, SMART achieves above $95\%$ success up to $50$ vehicles and remains feasible up to $90$ vehicles, with runtimes more than an order of magnitude faster than optimization-only approaches. Built on vehicle-to-everything communication, SMART incorporates vehicle-infrastructure cooperation through roadside sensing and agent coordination, improving scalability and safety. Real-world experiments further validate this design, achieving planning times as low as $0.014$\,s while preserving cooperative behaviors.

\end{abstract}

\begin{keyword}
Multi-vehicle trajectory planning, Behavior-level search, Convex corridor construction, Dense traffic environments 
\end{keyword}

\end{frontmatter}

%
\section{Introduction}

Coordinating large fleets of autonomous vehicles in dense and constrained environments poses a fundamental challenge for intelligent transportation systems. Recent advances in vehicle-to-everything (V2X) communication, particularly vehicle-to-infrastructure (V2I), enable richer information exchange among vehicles and roadside units, providing a foundation for large-scale cooperative driving. Within such connected environments, the task of multi-vehicle trajectory planning (MVTP) requires computing collision-free and kinematically feasible trajectories for all vehicles to reach destinations efficiently~\citep{zhang_multi-uncertainty_2024,huang_lead_2025}. However, as the number of vehicles increases, pairwise collision constraints grow quadratically, intensifying the problem’s non-convexity and often leading to infeasibility in confined spaces. A key to solving MVTP lies in exploring diverse homotopy classes, distinct interaction patterns that determine whether vehicles can resolve conflicts and complete tasks~\citep{dayan_near-optimal_2023,theurkauf_chance-constrained_2024}.

To address this, various planning strategies have been proposed, including coupled optimization~\citep{nascimento_multi-robot_2016,schouwenaars_mixed_2001}, distributed planning~\citep{rey_fully_2018,tordesillas_mader_2022}, sampling-based methods~\citep{lukyanenko_probabilistic_2023,shome_drrt_2020}, constraint-relaxation methods~\citep{ouyang_fast_2022,li_optimal_2021}, convex corridor construction~\citep{shi_neural-swarm2_2022,park_efficient_2020}, and grid-based search~\citep{park_online_2022, lee_dmvc-tracker_2025}. Coupled methods ensure optimality but quickly become computationally intractable as vehicle density increases~\citep{wang_safety_2016,nikou_scalable_2020}. Distributed and sampling-based approaches improve scalability but often fail in congested scenarios due to deadlocks or poor coordination~\citep{senbaslar_robust_2019, reis_control_2021}. Corridor-based methods depend heavily on high-quality initial guesses and can only explore a limited set of behaviors~\citep{ shome_drrt_2020}. Grid-based planning methods inspired by multi-agent pathfinding (MAPF) can efficiently explore interaction modes, but they face trade-offs between spatial resolution and feasibility~\citep{ wen_cl-mapf_2022,meng2025advances}. Overall, existing approaches either fail to capture the global interaction structure or incur prohibitive computation, limiting their applicability to real-time dense scenarios.

To overcome these limitations, we introduce SMART, Scalable Multi-Agent Reasoning and Trajectory Planning, a hierarchical framework designed for efficient and feasible MVTP in dense environments. SMART decouples behavior-level reasoning from trajectory optimization: a centralized priority-based search module first explores high-level homotopy classes using large-step reasoning guided by learned heuristics~\citep{yang2024csdo,huang_general_2024}, and a decentralized convex optimization module then resolves local conflicts by refining trajectories in parallel. Extensive simulations and real-world experiments demonstrate that SMART achieves higher success rates and faster computation compared to state-of-the-art methods, particularly in large-scale and high-density settings. The contributions are as follows:

\begin{itemize}
    \item We propose SMART, a scalable multi-agent planning framework with a hierarchical architecture. The upper layer uses large-step grid search to efficiently explore high-level interaction modes, while the lower layer refines trajectories with small-step optimization. This design addresses limitations in homotopy coverage and exploration efficiency.

    \item We introduce a parallelizable optimization method based on corridor construction, which decouples the joint non-convex problem into independent convex subproblems. By leveraging initial guesses, the method efficiently generates robust feasible corridors and solves each vehicle’s trajectory in parallel.


    \item We validate SMART in fully connected environments, showing strong scalability and real-time performance. On $50$\,m $\times$ $50$\,m maps, it sustains over $90\%$ success within $1$\,s up to $25$ vehicles, and on $100$\,m $\times$ $100$\,m maps achieves above $95\%$ up to $50$ and remains feasible at $90$. With V2I support, roadside sensing and coordination enhance multi-agent interaction. Real-world tests confirm practicality, with planning times as low as $0.014$\,s while preserving cooperative behaviors.

\end{itemize}

The remainder of this paper is organized as follows. Section 2 reviews related works. Section 3 presents the methodology, including behavior search and trajectory optimization. Section 4 reports experimental results in simulation and real-world tests. Section 5 concludes the paper and discusses future directions.

%
\section{Related Works}
\label{rw}
%
Existing approaches to multi-vehicle trajectory planning (MVTP) can be broadly categorized into six classes: coupled optimization, distributed planning, sampling-based planning, constraint-relaxation methods, convex corridor construction, and grid-based search. While each class offers valuable contributions, face limitations in scalability, completeness, or robustness when applied to dense environments~\citep{nascimento_multi-robot_2016,ouyang2022fast,yang2023decoupled}. 

Coupled optimization methods formulate MVTP as a unified high-dimensional control problem by treating all vehicle trajectories as a single decision variable~\citep{nascimento_multi-robot_2016, mellinger_minimum_2011}. These methods provide theoretical guarantees of completeness and optimality. For example, early works modeled MVTP as mixed-integer quadratic or linear programming problems by transforming pairwise collision constraints into linear inequalities~\citep{schouwenaars_mixed_2001-1}. Although theoretically rigorous, these methods are computationally intensive. Even in simplified settings, solving such formulations requires tens or even hundreds of seconds, making them impractical for real-time applications. As the number of vehicles increases, the number of pairwise constraints grows quadratically, and the optimization time escalates sharply, severely limiting scalability.

Distributed planning approaches treat each vehicle as an independent agent, solving local collision avoidance iteratively. Techniques based on model predictive control and control barrier functions (CBFs) have shown promising results~\citep{ senbaslar_robust_2019, rey_fully_2018}. For instance, CBF-based quadratic programs can solve velocity commands for hundreds of robots within milliseconds ~\citep{reis_control_2021, allibhoy_control-barrier-function-based_2024}. However, these methods often assume simple dynamics and require conservative parameter tuning. In dense environments, distributed planners tend to suffer from deadlocks or livelocks due to a lack of global coordination, and their performance degrades rapidly with increasing agent density.

Sampling-based methods extend single-agent motion planning algorithms, such as probabilistic roadmaps (PRM) and rapidly-exploring random trees (RRT), to multi-vehicle systems ~\citep{lukyanenko_probabilistic_2023,shome_drrt_2020}. These methods are probabilistically complete and can approach optimality as the number of samples increases. However, their effectiveness declines in high-density scenarios due to poor coordination and excessive sampling demands. Even in small indoor setups with five vehicles, planning can take several seconds, and scaling to ten or more agents remains challenging~\citep{dayan_near-optimal_2023}.

Constraint-relaxation methods attempt to simplify non-convex optimization by iteratively linearizing difficult constraints around previous solutions~\citep{li_optimal_2021, ma_lifelong_2017, benedikter2019convex}. Sequential convex programming (SCP) and incremental variants such as iSCP aim to improve convergence and handle infeasibility by selectively relaxing collision constraints~\citep{marcucci2024graphs,chen2015decoupled}. Although such methods can generate feasible trajectories in new homotopy classes, they still require solving non-convex problems and often remain sensitive to initial guesses. Their computational cost remains high for large-scale MVTP problems~\citep{ouyang_fast_2022}.

Corridor-based methods construct convex safe corridors around reference trajectories to transform inter-vehicle and obstacle avoidance constraints into convex formulations~\citep{park_efficient_2020-1, shi_neural-swarm2_2022}. This strategy ensures separation among vehicles by assigning distinct corridors to each agent. Some implementations generate corridors using MAPF solutions as initial guesses and have demonstrated reasonable performance in sparse warehouse environments~\citep{park_online_2022,lee_dmvc-tracker_2025}. However, because corridors are constructed around fixed reference points, these methods tend to explore only a limited set of homotopy classes and may fail without high-quality initial paths.

Grid-based search methods discretize state, space, and motion into a finite graph, enabling the planner to reason over high-level interaction modes between vehicles~\citep{zhang_d-pbs_2024,lee_dmvc-tracker_2025}. Inspired by MAPF, these approaches can efficiently explore a wide range of homotopy classes. Frameworks such as CL-MAPF have shown success in adapting hybrid A* methods to nonholonomic multi-agent systems~\citep{wen_cl-mapf_2022}, and enhancements like priority-based search and dual-channel expansions improve success rates~\citep{lin_multi-agent_2025}. However, the trade-off between resolution and feasibility remains a key challenge: small step sizes incur high computational cost, while large steps may result in dynamic infeasibility or collisions.

While each of these methods contributes toward solving MVTP, they either fail to explore diverse interaction modes efficiently or cannot scale to real-time performance in dense multi-agent settings. A robust solution must unify global behavior-level reasoning with tractable local trajectory refinement while preserving feasibility, efficiency, and coordination.

%
\section{Methodology}
\label{methodology}
%

\subsection{MVTP: Problem Formulation}\label{sec:mvtp}

\begin{figure}[htbp]
\centering
\includegraphics[width=0.85\textwidth]{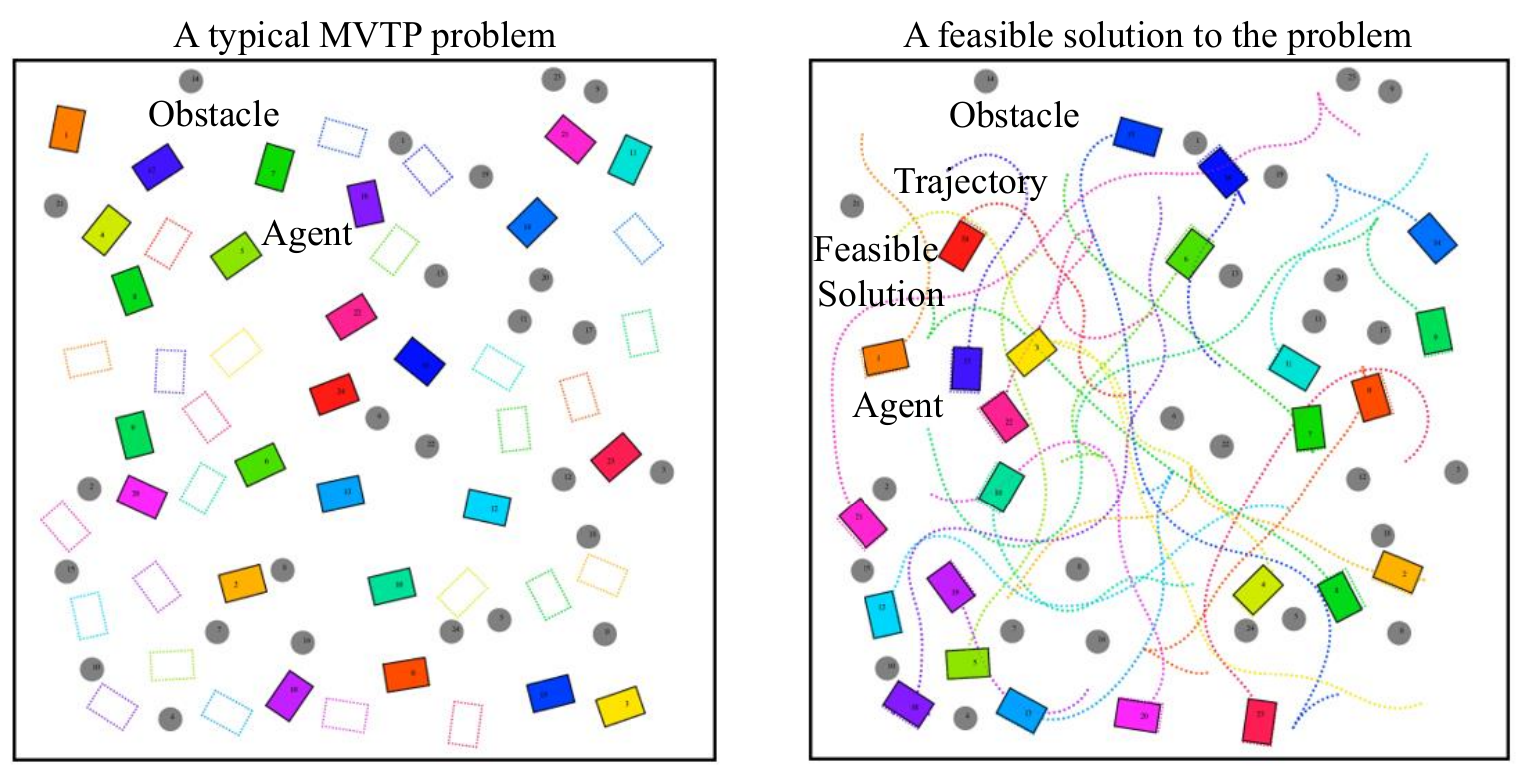}
\caption{A typical MVTP instance and one feasible solution.
(a) A typical MVTP problem specifying map size, static obstacles, and vehicle start/goal configurations. 
(b) A corresponding feasible solution where all vehicles reach their goals without collisions.}
\label{fig:fig1}
\end{figure}

\textbf{Definition 1 (Multi-Vehicle Trajectory Planning, MVTP).} A typical MVTP instance and one feasible solution are illustrated in Fig.~\ref{fig:fig1}.
Given the map size, static obstacles, and for each vehicle the start and goal configurations
(position and heading), the task is to compute collision-free, kinematically feasible, and
time-optimal trajectories for all vehicles.

We consider a system of $M$ front-wheel–steering vehicles
$\{a^{(1)},a^{(2)},\dots,a^{(M)}\}$ that operate in a continuous planar workspace
$\mathcal{W}\subset\mathbb{R}^2$, where a subset $\mathcal{O}\subset\mathcal{W}$ is
occupied by static obstacles. For simplicity, let $[M]=\{1,2,\dots,M\}$ denote the
set of vehicle indices, and use superscript $(i)$ for quantities of vehicle $a^{(i)}$.

The state of each vehicle is
\begin{equation}
z = [x,\, y,\, \theta,\, \phi]^\top \in \mathbb{R}^4,
\end{equation}
where $(x,y)$ is the rear-axle center, $\theta$ is the yaw angle, and $\phi$ is the steering angle.
The control input is
\begin{equation}
u = [v,\, \omega]^\top \in \mathbb{R}^2,
\end{equation}
where $v$ is the longitudinal speed and $\omega$ is the steering rate.

The trajectory of vehicle $a^{(i)}$ is represented as
\begin{equation}
\mathcal{T}^{(i)} = \Big[z^{(i)}_0, z^{(i)}_1, \dots, z^{(i)}_{\tau_f^{(i)}} \Big], 
\label{eq:trajectory}
\end{equation}
where $\tau_f^{(i)}+1$ is the number of sampled states. Each trajectory must begin at
the initial state $s^{(i)}$ and terminate at the goal state $g^{(i)}$, i.e.,
\begin{equation}
z^{(i)}_0 = s^{(i)}, \quad z^{(i)}_{\tau_f^{(i)}} = g^{(i)}, \quad \forall i \in [M].
\label{eq:boundary}
\end{equation}

The arrival time of vehicle $a^{(i)}$ is denoted by $\tau_f^{(i)}$, and the overall
mission completion time is defined as
\begin{equation}
\tau_f = \max_{i \in [M]} \, \tau_f^{(i)}.
\label{eq:completion}
\end{equation}

Following standard MAPF assumptions, once a vehicle reaches its goal it remains there
for all subsequent time steps:
\begin{equation}
z^{(i)}_{t} = g^{(i)}, \quad \forall \tau_f^{(i)} \le t \le \tau_f,\ \forall i \in [M].
\label{eq:stay_goal}
\end{equation}

To explicitly describe collision-avoidance requirements, we define the occupied region
of a vehicle at state $z$ as $\mathcal{R}(z)\subseteq \mathcal{W}$. A valid MVTP solution
must guarantee that vehicles do not overlap at their initial or goal configurations:
\begin{equation}
\mathcal{R}(s^{(i)}) \cap \mathcal{R}(s^{(j)}) = \emptyset, 
\quad \forall i,j \in [M], \, i \neq j,
\label{eq:init_collision}
\end{equation}
\begin{equation}
\mathcal{R}(g^{(i)}) \cap \mathcal{R}(g^{(j)}) = \emptyset, 
\quad \forall i,j \in [M], \, i \neq j.
\label{eq:goal_collision}
\end{equation}

Furthermore, at every time step, each vehicle must avoid both static obstacles and other
vehicles. Formally, these constraints are expressed as
\begin{equation}
\mathcal{R}(z^{(i)}_t) \cap \mathcal{O} = \emptyset, 
\quad \forall t \geq 0, \, \forall i \in [M],
\label{eq:static_collision}
\end{equation}
\begin{equation}
\mathcal{R}(z^{(i)}_t) \cap \mathcal{R}(z^{(j)}_t) = \emptyset,
\quad \forall t \geq 0, \, \forall i,j \in [M], \, i \neq j.
\label{eq:vehicle_collision}
\end{equation}

In summary, the solution $X$ is defined as the set of dynamically feasible and
collision-free trajectories for all vehicles. Its quality is measured by the task
completion time $\tau_f$. The MVTP problem can thus be formulated as the following
optimal control program:
\begin{equation}
\text{Minimize} \quad \tau_f,
\label{eq:objective}
\end{equation}
subject to the boundary conditions~\eqref{eq:boundary}, vehicle dynamics, static obstacle
constraints~\eqref{eq:static_collision}, and inter-vehicle collision-avoidance
constraints~\eqref{eq:vehicle_collision}.
This formalization highlights the intrinsic non-convexity of MVTP and provides the
foundation for the hierarchical planning and optimization framework.


\subsection{Front-End Centralized Behavior Search} 

\subsubsection{High-level priority-based search}

At the upper layer, we adopt Priority-Based Search (PBS) to explore feasible interaction
modes. The key idea is that if a collision occurs in the solution of a node, PBS resolves it
by branching on the relative order of the colliding vehicles. For each conflict, two child
nodes are generated by adding opposite priority constraints between the pair of vehicles,
ensuring that alternative resolution options are systematically considered. The process
continues until all conflicts are eliminated or no feasible plan exists.

\begin{algorithm}[htbp]
\caption{Priority-Based Search with Warm-Start and Selective Replanning}
\label{alg:pbs}
\KwIn{MAPF instance $(\mathcal{O}, s, g)$}
\KwOut{Feasible solution or failure}
\BlankLine
\tcp{Warm-start root generation}
$\ell_{\text{APRIL}} \gets S2AN()$; \\
Initialize root node $v_{\text{Root}}$ with empty constraints; \\
Initialize higher-priority trajectory set $\mathcal{T}_h \gets \emptyset$; \\
\ForEach{$i \in \ell_{\text{APRIL}}$}{
    $\mathcal{T}^{(i)} \gets \text{STHA}^*(\mathcal{O}, \mathcal{T}_h, s^{(i)}, g^{(i)})$; \\
    \If{$\mathcal{T}^{(i)} = \emptyset$}{
        Attempt independent planning for $a^{(i)}$; \\
        \If{$\mathcal{T}^{(i)} = \emptyset$}{ \Return failure }
    }
    $\mathcal{T}_h \gets \mathcal{T}_h \cup \{\mathcal{T}^{(i)}\}$; \\
}
$v_{\text{Root}}.\mathcal{T} \gets \{\mathcal{T}^{(i)} \mid i \in [M]\}$; \\
\BlankLine
\tcp{Node update under new priority constraints}
\ForEach{new constraint $(a^{(i)}, a^{(j)})$ added to node $N'$}{
    $\ell \gets \text{TopologicalSort}(\{a^{(i)}\}\cup\{a^{(k)} \mid a^{(i)} \prec_{N'} a^{(k)}\})$; \\
    $\mathcal{T}_h \gets \{\mathcal{T}^{(p)} \mid a^{(p)} \prec_{N'} a^{(i)}\}$; \\
    \ForEach{$a^{(m)} \in \ell$}{
        \If{$a^{(m)} = a^{(i)}$ \textbf{or} $a^{(m)}$ collides with higher-priority vehicles}{
            $\mathcal{T}^{(m)} \gets \text{STHA}^*(\mathcal{O}, \mathcal{T}_h, s^{(m)}, g^{(m)})$; \\
            \If{$\mathcal{T}^{(m)} = \emptyset$}{ mark $N'$ infeasible; \Return failure }
            $N'.\mathcal{T}[m] \gets \mathcal{T}^{(m)}$; \\
            $\mathcal{T}_h \gets \mathcal{T}_h \cup \{\mathcal{T}^{(m)}\}$; \\
        }
    }
}
\Return feasible solution if all conflicts are resolved;
\end{algorithm}

The root node is initialized with an empty set of constraints, meaning all vehicles are 
planned independently. If the solution is collision-free, it is valid for the MVTP instance. 
Otherwise, at least one vehicle pair conflicts at some time step. One such conflict is 
selected, and the two possible priority orders generate child nodes, each requiring 
replanning under the new constraints. Nodes yielding infeasible plans are discarded, while 
feasible ones are further expanded. A depth-first traversal of the priority tree continues 
until a conflict-free solution is found or the search space fully explored. In this way, PBS 
systematically explores alternative interaction patterns and provides structured initial 
trajectories for subsequent low-level refinement.

To accelerate the search, two techniques are used: (1) \textit{warm-start initialization}, and
(2) \textit{selective replanning}. For warm-start, we leverage the Synthetic Score-based Attention Network (S2AN)~\cite{yang2024attention}, 
a model proposed in our prior work. 
S2AN takes as input a multi-agent pathfinding instance with agent start/goal states and static obstacles, 
and outputs a heuristic global priority order $\ell_{\text{APRIL}}$. This order reflects learned attention-based 
scores over potential priority sequences, thereby guiding the initial planning order. Although not enforced 
as hard constraints to preserve completeness, $\ell_{\text{APRIL}}$ is used as the initial sequence for planning. 
Vehicles are planned accordingly, and if one fails due to blocking by higher-priority vehicles, the algorithm 
relaxes dependencies and replans independently. If it still fails, the root is declared infeasible; otherwise, 
the root node contains all planned trajectories, initializing the PBS tree with a high-quality starting point 
for branching.

\begin{figure}[htbp]
\centering
\includegraphics[width=0.95\textwidth]{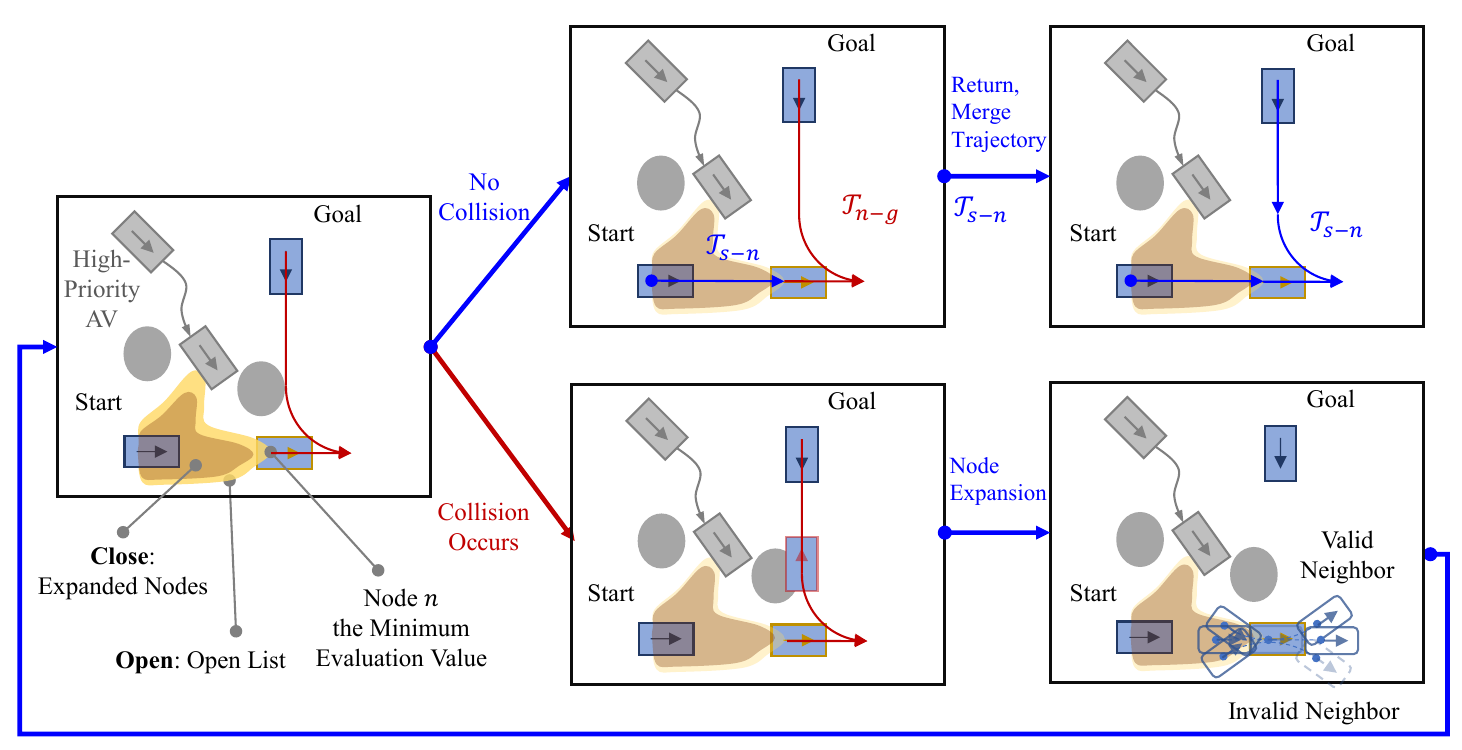}
\caption{Spatiotemporal Hybrid A* (STHA*) algorithm illustration. 
Nodes are expanded in the joint space-time domain according to the minimum 
evaluation function $f = g + h$. When a node reaches the goal without collision, 
a feasible trajectory is reconstructed by backtracking. If a collision occurs, the search 
branches on alternative priority orders, and invalid neighbors are pruned. Valid neighbors 
are propagated toward the goal through node expansion, ensuring dynamic feasibility.}
\label{fig:fig2}
\end{figure}

When new priority constraints are introduced, the affected node must be updated by
replanning for the low-priority vehicle and its dependent set. A topological sort of the
partial order is performed to ensure consistency and generate a valid total order. Only
vehicles directly involved in the new constraint or colliding with higher-priority vehicles
are replanned, which significantly reduces redundant computation. If any vehicle fails to
find a valid trajectory under these updated constraints, the node is marked infeasible and
discarded; otherwise, the updated node remains valid and continues to expand in the tree.
The overall procedure, combining warm-start root generation with selective replanning,
is summarized in Algorithm~\ref{alg:pbs}.


\subsubsection{Low-level single-vehicle trajectory planning}

At the low level, each vehicle computes its individual trajectory using the 
Spatiotemporal Hybrid A* (STHA*) algorithm. The inputs include the static
obstacle set $\mathcal{O}$, the trajectories of higher-priority vehicles
$\{\mathcal{T}^{(j)} \mid a^{(j)} \prec_N a^{(i)}\}$, and the start and goal
states $s^{(i)}$ and $g^{(i)}$. The algorithm outputs a dynamically feasible
trajectory for vehicle $a^{(i)}$ under Ackermann steering constraints. 
STHA* guarantees completeness and optimality: if a feasible solution exists, 
it will be found, and if the planner fails, this implies that no trajectory 
can reach the goal without collisions.

\begin{algorithm}[htbp]
\caption{Spatio-Temporal Hybrid A* (STHA*) for Single-Vehicle Planning}
\label{alg:stha}
\KwIn{Static obstacles $\mathcal{O}$, higher-priority trajectories $\mathcal{T}_h$, start $s^{(i)}$, goal $g^{(i)}$}
\KwOut{Feasible trajectory $\mathcal{T}^{(i)}$ or failure}
\BlankLine
Initialize $\mathcal{V}_{\text{open}} \gets \{(s^{(i)}, f=0)\}$, $\mathcal{V}_{\text{close}} \gets \emptyset$; \\
\While{$\mathcal{V}_{\text{open}} \neq \emptyset$}{
    $n \gets$ node in $\mathcal{V}_{\text{open}}$ with smallest $f$; \\
    Move $n$ to $\mathcal{V}_{\text{close}}$; \\
    \If{$n = g^{(i)}$ and collision-free}{
        Construct path $\mathcal{T}^{(i)}$ by backtracking; \\
        \Return $\mathcal{T}^{(i)}$;
    }
    \ForEach{action $a \in \mathcal{A}_{\text{actions}}$}{
        Generate successor $n'$ from $n$ under $a$; \\
        \If{$n'$ collides with $\mathcal{T}_h$ or $\mathcal{O}$}{ continue; }
        \If{$n' \in \mathcal{V}_{\text{close}}$}{ continue; }
        Compute $g(n') = g(n) + c(a)$ and $f(n') = g(n') + h(n', g^{(i)})$; \\
        Record parent of $n'$ as $n$; \\
        \If{$n' \notin \mathcal{V}_{\text{open}}$}{ insert $n'$ into $\mathcal{V}_{\text{open}}$; }
        \Else{ update $n'$ in $\mathcal{V}_{\text{open}}$ if lower cost found; }
    }
}
\Return failure;
\end{algorithm}

Compared with the standard Hybrid A*, STHA* augments the search space
with the temporal dimension to explicitly handle dynamic collision-avoidance.
As illustrated in Fig.~\ref{fig:fig2}, the search expands nodes in the joint
state-time domain by repeatedly selecting the node with the smallest evaluation
function $f = g + h$, where $g$ is the accumulated path cost and $h$ is a
heuristic estimate of the remaining cost-to-go. Candidate nodes are maintained
in a priority queue $\mathcal{V}_{\text{open}}$, while expanded nodes are
stored in a hash table $\mathcal{V}_{\text{close}}$. At each step, if the
current node reaches the goal without collision, the solution trajectory is
reconstructed by backtracking. Otherwise, all valid neighbors are generated
and inserted into the open list, while invalid neighbors are pruned. This process
continues until either a feasible trajectory is returned or the open list is
exhausted. The detailed procedure of the spatiotemporal Hybrid A* algorithm is given in Algorithm~\ref{alg:stha}.

%

\subsection{Back-End Distributed Optimization}

The back-end module adopts a distributed sequential quadratic programming (SQP)
framework. Starting from an initial trajectory guess, the method interpolates discrete
states, partitions inter-vehicle collision constraints via neighborhood search, constructs
safe spatio-temporal corridors to encode static obstacles, and finally solves a set of
quadratic programs (QPs) iteratively. Each vehicle updates its trajectory within its own
corridor, enabling parallel optimization across the fleet. This distributed design not only
reduces the dimensionality of the joint optimization problem but also ensures scalability
to large numbers of vehicles, while maintaining feasibility and smoothness of the planned
trajectories.

\subsubsection{Interpolation of the initial guess}

The input is the initial guess $\mathbf{X}_{\text{search}}$, and the output is the interpolated
trajectory $\bar{\mathbf{X}}_{0}$. The overline indicates constant values associated with
the initial guess, while the subscript $0$ denotes the first iteration.
For example, $\theta^{(i)}_{t,0}$ is the interpolated yaw angle of vehicle $a^{(i)}$ at time $t$.

Interpolation is performed by inserting $n_{\text{interp}}$ points along each segment of
the Reed-Shepp path that composes the initial guess. The time step between adjacent
points is computed as
\begin{equation}
\Delta t = \frac{s_{\text{action}}}{(n_{\text{interp}}+1)\,v_{\max}},
\label{eq:interp}
\end{equation}
where $s_{\text{action}}$ is the distance of the motion primitive and $v_{\max}$ is the
maximum velocity. Since the Reed-Shepp trajectory encodes vehicle orientation, the
interpolation also provides heading $\theta$ and related control quantities such as
velocity and steering rate. The sequence forms the fixed initial trajectory
$\bar{\mathbf{X}}_{0}$ for optimization.

In practice, the interpolated guess may contain minor conflicts, including:
(A) collisions between vehicles, (B) trajectories crossing workspace boundaries,
and (C) collisions with static obstacles. These conflicts are subsequently handled
through safe corridor construction and distributed optimization.

\subsubsection{Neighborhood set search}

Given the interpolated initial guess $\bar{\mathbf{X}}_{0}$, the purpose of neighborhood search is to construct the set $\mathcal{B}_{\text{neighbor}}$, which is later used for collision detection and distance evaluation in distributed optimization.

\begin{figure}[htbp]
\centering
\includegraphics[width=1\textwidth]{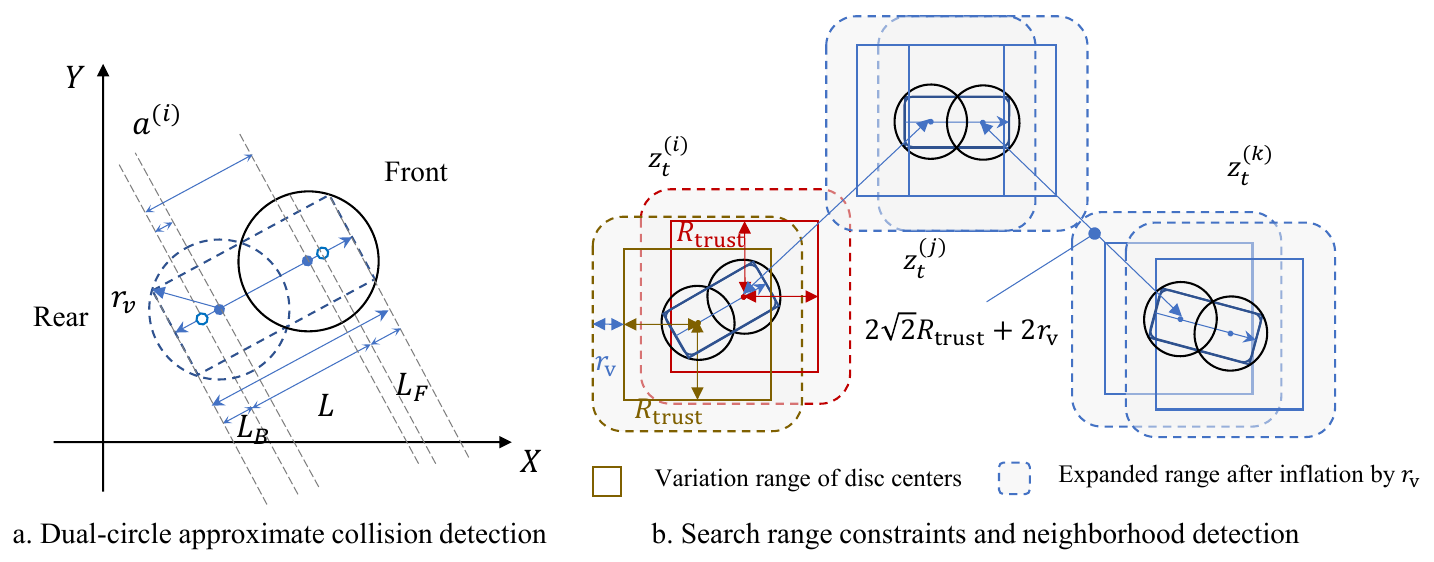}
\caption{Dual-circle approximation and neighborhood detection. 
(a) Each vehicle is represented by two discs of radius $r_v$, whose motion defines the trust region $R_{\text{trust}}$ around the interpolated trajectory. 
(b) Neighborhood detection based on inflated trust regions: overlapping regions imply potential collisions, while sufficiently distant vehicles are excluded as non-neighbors.}
\label{fig:fig3}
\end{figure}

As illustrated in Fig.~\ref{fig:fig3}(a), each vehicle is approximated by two discs of radius $r_v$, placed at the quarter points along the longitudinal axis. This dual-circle representation captures the vehicle geometry under Ackermann steering while allowing analytical computation of inter-vehicle distances. The trajectories of these disc centers define the trust region $R_{\text{trust}}$, which restricts the deviation of the optimized trajectory from the interpolated initial guess. The radius $r_v$ is chosen such that the two discs completely cover the vehicle body.

Fig.~\ref{fig:fig3}(b) shows how these trust regions are further used for neighborhood detection. Specifically, each trust region is inflated by $r_v$. If two inflated regions overlap, the corresponding vehicles are identified as neighbors, and collision constraints are enforced in the optimization. Conversely, if the Euclidean distance between two vehicles exceeds $2\sqrt{2}\,R_{\text{trust}}+2r_v$, they are guaranteed to be non-neighbors and can be excluded from pairwise checks. This pruning strategy eliminates redundant collision constraints and significantly accelerates the distributed optimization procedure.

Formally, let $\mathbf{Y}^F_t$ and $\mathbf{Y}^R_t$ denote the front and rear circle centers of vehicle $a^{(i)}$ at time $t$. The pairwise distance between two vehicles $a^{(i)}$ and $a^{(j)}$ is defined as
\begin{equation}
f_{\text{dist}}\big(z^{(i)}_t, z^{(j)}_t\big) =
\min \Big( \|\mathbf{Y}^F_t{}^{(i)} - \mathbf{Y}^F_t{}^{(j)}\|,\;
           \|\mathbf{Y}^F_t{}^{(i)} - \mathbf{Y}^R_t{}^{(j)}\|,\;
           \|\mathbf{Y}^R_t{}^{(i)} - \mathbf{Y}^F_t{}^{(j)}\|,\;
           \|\mathbf{Y}^R_t{}^{(i)} - \mathbf{Y}^R_t{}^{(j)}\| \Big) - 2r_v,
\label{eq:fdist}
\end{equation}
where a collision occurs if $f_{\text{dist}}(z^{(i)}_t, z^{(j)}_t) \leq 0$.

In addition, to ensure that optimized trajectories remain consistent with the initial guess, a trust region constraint is imposed on each circle center:
\begin{equation}
\big\| \mathbf{Y}_{c,t}^{(i)} - \bar{\mathbf{Y}}_{c,t}^{(i)} \big\|
\leq R_{\text{trust}}, \quad \forall i \in [M],\; 0 \leq t \leq \tau_f,
\label{eq:trust}
\end{equation}
where $\mathbf{Y}_{c,t}^{(i)}$ denotes the circle center of vehicle $a^{(i)}$ at time $t$, and $\bar{\mathbf{Y}}_{c,t}^{(i)}$ is the corresponding interpolated center. This constraint narrows the search space and prevents excessive deviation from the nominal trajectory.

\subsubsection{Robust corridor generation and SQP solving}

To model static collision avoidance, we adopt a corridor-based representation.
Collisions are expressed via Minkowski difference: ensuring that a vehicle’s
circle of radius $r_v$ remains within the map is equivalent to eroding the
workspace boundary by $r_v$. In other words, the feasible region is shrunk
such that the circle can move without exceeding the boundary.

Starting from an initial safe point, feasible corridors are constructed by expanding
rectangular boxes along four directions until encountering an obstacle, an eroded
boundary, or the maximum allowable range. Each feasible box defines a local region
guaranteed to be free of static collisions. By sweeping along the trajectory points,
a sequence of feasible boxes can be generated, forming a safe corridor. This process
is robust even when the interpolated guess lies slightly in collision, since the initial
point is shifted to the nearest safe position before box expansion.

The safe corridor constraints can be written in linear form as
\begin{equation}
\bar{\mathbf{A}}_{c,0}\,\mathbf{Y}_c \leq \bar{\mathbf{b}}_c,
\label{eq:linear_corridor}
\end{equation}
where $\bar{\mathbf{A}}_{c,0}$ and $\bar{\mathbf{b}}_c$ are constant matrices determined
by the separating hyperplanes. Specifically,
\begin{equation}
\bar{\mathbf{Y}}^{(i)}_{c,\min,t,k} \leq
\bar{\mathbf{A}}^{(i)}_{\text{static},t,k}\,\mathbf{Y}^{(i)}_{c,t,k+1}
\leq \bar{\mathbf{Y}}^{(i)}_{c,\max,t,k}, \quad
\forall i \in [M], \; 0 \leq t \leq \tau_f,
\label{eq:static_box}
\end{equation}
where $\bar{\mathbf{A}}^{(i)}_{\text{static},t,k}$ encodes the box normals,
and $\bar{\mathbf{Y}}^{(i)}_{c,\min,t,k}$, $\bar{\mathbf{Y}}^{(i)}_{c,\max,t,k}$
denote the box bounds in each coordinate.

The optimization objective is to smooth trajectories by minimizing variations in
velocity and steering rate:
\begin{equation}
J = \sum_t \left( \alpha_v \big(\Delta v^{(i)}_{t,k+1}\big)^2 +
                  \alpha_\omega \big(\omega^{(i)}_{t,k+1}\big)^2 \right),
\label{eq:objective_corridor}
\end{equation}
where $\alpha_v$ and $\alpha_\omega$ are penalty weights.

To enforce feasibility, the kinematic constraints are linearized around the previous
solution $\mathbf{X}^{(i)}_k$. Nonlinear mappings between vehicle states $z$ and
circle centers $\mathbf{Y}_c$ are also linearized. The resulting sequential quadratic
program (SQP) for vehicle $a^{(i)}$ is
\begin{align}
\min_{\mathbf{X}^{(i)}} \quad & J \label{eq:sqp}\\
\text{s.t.} \quad &
z^{(i)}_{0,k+1} = s^{(i)}, \quad z^{(i)}_{\tau_f,k+1} = g^{(i)}, \nonumber\\
& |v^{(i)}_{t,k+1}| \leq v_{\max}, \;
  |\omega^{(i)}_{t,k+1}| \leq \omega_{\max}, \quad \forall t < \tau_f, \nonumber\\
& |\phi^{(i)}_{t,k+1}| \leq \phi_{\max}, \quad \forall t \leq \tau_f, \nonumber\\
& \text{Corridor constraints~\eqref{eq:linear_corridor}--\eqref{eq:static_box}}. \nonumber
\end{align}

By iteratively solving the above SQPs for all vehicles, new solutions
$\mathbf{X}_{k+1}$ are generated. The process terminates when the difference
$\|\mathbf{X}_{k+1}-\mathbf{X}_k\|_2$ falls below a threshold, or when the
linearized solution is feasible in the original nonlinear problem. The final result
is a smooth, collision-free set of trajectories.

%
\section{Experiments}

To evaluate the effectiveness of the proposed method, we conduct experiments on
three types of environments: obstacle-free maps, randomly generated obstacle maps,
and indoor maps. The number of vehicles is gradually increased in each scenario to
create progressively denser environments, thereby testing the scalability and efficiency
of the method for large-scale multi-vehicle trajectory planning.

\subsection{Simulation Design and Analysis}

All experiments are implemented in C++ and executed on an Intel Xeon Gold 622R
CPU at 2.90~GHz. To ensure fairness, FOTP is executed in Matlab on an Intel Core
i7-9750H CPU at 2.26~GHz, and minor normalization is applied to account for 
programming language and hardware differences. Performance is evaluated by 
progressively increasing the number of vehicles and measuring the ability of each 
method to compute feasible trajectories in dense environments.

\subsubsection{Simulation Setup}

\begin{figure}[htbp]
\centering
\includegraphics[width=0.95\textwidth]{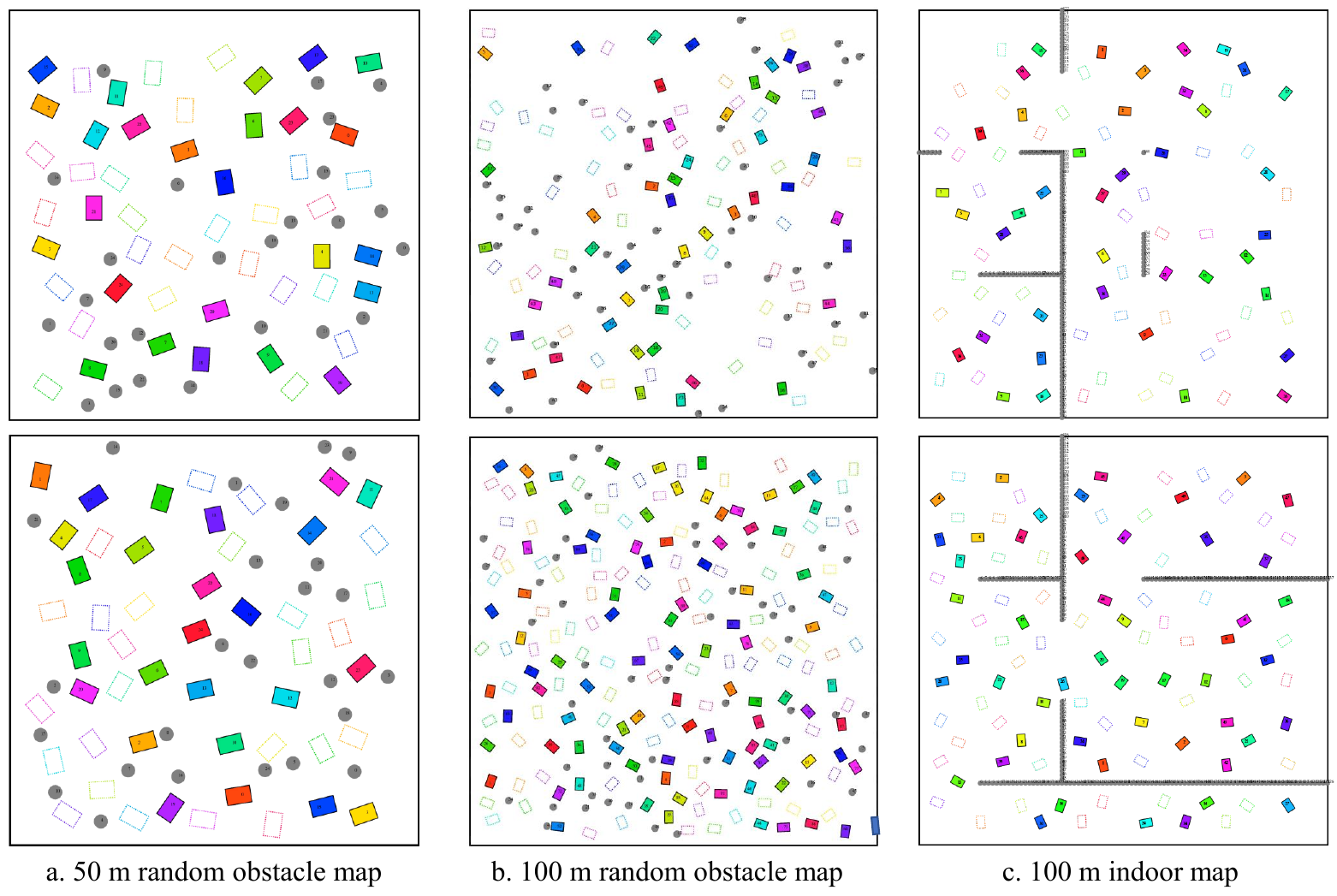}
\caption{Representative examples of simulation environments.}
\label{fig:fig4}
\end{figure}

The benchmark consists of two map sizes, $50 \times 50$\,m and $100 \times 100$\,m, 
each containing three environment types: obstacle-free layouts, randomly generated 
obstacle maps, and random indoor maps. The number of vehicles ranges from $5$ to $100$, 
with start and goal states uniformly sampled across the workspace. For each combination 
of map type and vehicle number, $60$ random instances are generated, resulting in a 
total of $2100$ test cases. Representative examples are shown in Fig.~\ref{fig:fig4}.

All vehicles are modeled as identical front-wheel–steering cars with realistic geometric 
and kinematic limits, ensuring consistency across different simulation settings. The 
minimum turning radius is $3$\,m, with a vehicle width of $1$\,m and a wheelbase of 
$2$\,m. Both the forward and rear overhangs are set to $1$\,m, and the maximum 
steering angle is constrained to $0.3218$\,rad, capturing the nonholonomic limitations 
of real vehicles. Dynamic bounds include a maximum speed of $2.0$\,m/s and a maximum 
steering rate of $0.07$\,s$^{-1}$, reflecting typical maneuverability in confined spaces. For the search layer, the primitive step length is $0.706$\,m, which balances resolution 
and computational efficiency. Penalty weights are assigned to steering, reversing, and 
heading changes to discourage inefficient maneuvers and promote natural driving 
behaviors. In the optimization layer, a trust region of $2.0$\,m is imposed to constrain 
trajectory deviations from the initial guess, thereby ensuring stability during iterative 
updates and improving convergence toward smooth, feasible solutions. This unified 
parameterization provides a fair and controlled basis for comparing different planning 
approaches under varied traffic densities and map complexities.

\subsubsection{Results and Analysis}

\begin{figure}[htbp]
\centering
\includegraphics[width=1\textwidth]{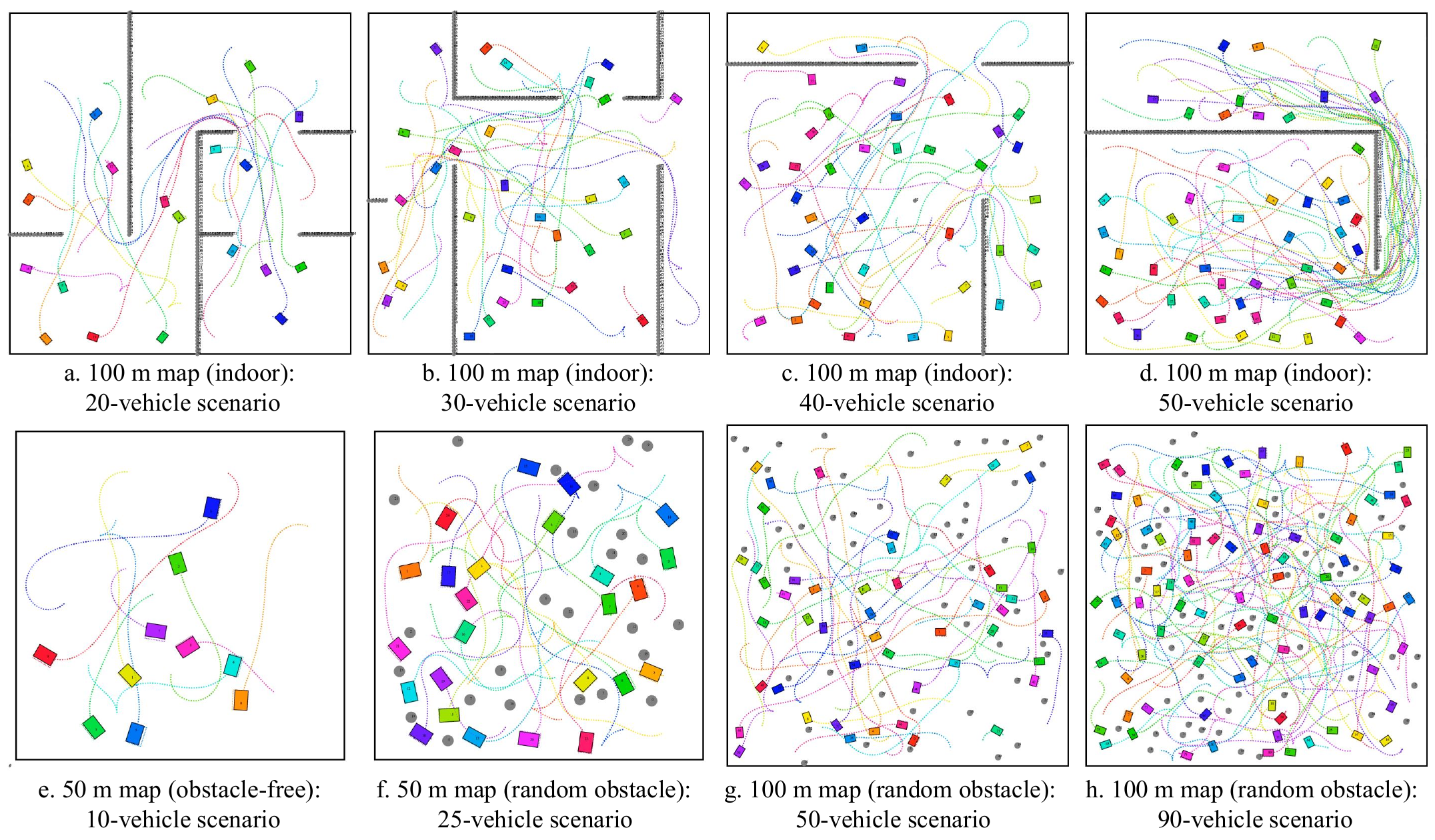}
\caption{Representative solved trajectories under different vehicle numbers.}
\label{fig:fig5}
\end{figure}

As illustrated in Fig.~\ref{fig:fig5}, the proposed method generates feasible 
and collision-free trajectories across a wide range of scenarios, including 
indoor layouts, obstacle-free maps, and random obstacle maps, under varying 
vehicle densities. Each subfigure shows the final positions of all vehicles at 
the end of execution, with gray areas denoting obstacles and dotted lines 
representing traveled paths. Even as the number of vehicles increases from 
tens to nearly one hundred, the method maintains feasibility. This is achieved 
by the PBS-based front-end, which provides high-quality initial solutions, and 
the distributed back-end optimization, which refines them into smooth, 
kinematically feasible trajectories.

To further assess performance, we compare our method with representative MVTP
approaches: Sequential Car-like Conflict search (SCC)~\citep{liu2024safer}, Priority Planning (PP)~\citep{vcap2015prioritized},
and Fast and Optimal Trajectory Planning (FOTP)~\citep{ouyang2022fast}. SCC extends CBS with a
car-like motion model and discretized state/action space, PP plans trajectories
based on randomized vehicle orderings, and FOTP represents optimization-based
planning baselines. Metrics include success rate, task completion time, and
computation time. The results are summarized in Tables~\ref{tab:full_comparison_eng} and \ref{tab:100m_combined_note}.

\begin{table}[htbp]
\centering
\footnotesize
\setlength{\tabcolsep}{4pt}            
\renewcommand{\arraystretch}{1.2}      
\caption{Comparison results on $50\,\text{m} \times 50\,\text{m}$ random obstacle maps, 
$50\,\text{m} \times 50\,\text{m}$ obstacle-free maps, and $100\,\text{m} \times 100\,\text{m}$ indoor maps. 
Metrics include success rate ($\uparrow$), solution time/s ($\downarrow$), and task completion time/s ($\downarrow$).}
\label{tab:full_comparison_eng}
\begin{adjustbox}{width=\linewidth}
\begin{tabular}{c|cccc|cccc|cccc}
\hline
\multicolumn{13}{c}{\textbf{50 m Random Obstacle Maps}}\\
\hline
\multirow{2}{*}{Method} & \multicolumn{4}{c|}{Success Rate $\uparrow$} & \multicolumn{4}{c|}{Solution Time/s $\downarrow$} & \multicolumn{4}{c}{Task Completion Time/s $\downarrow$}\\
& 10 & 15 & 20 & 25 & 10 & 15 & 20 & 25 & 10 & 15 & 20 & 25 \\
\hline
SCC   & 95.00\% & 81.67\% & 36.67\% & 05.00\% & 0.42 & 2.16 & 7.14 & 13.39 & 48.74 & 52.21 & 54.99 & 54.55 \\
PP    & 95.00\% & 83.33\% & 85.00\% & 51.67\% & 0.14 & 0.36 & 2.13 & 2.80  & 49.70 & 53.29 & 60.03 & 65.17 \\
FOTP  & 73.33\% & 70.00\% & 65.00\% & 51.67\% & 4.90 & 6.12 & 50.64 & 90.00 & 61.58 & 67.87 & 73.11 & 73.11 \\
SMART & \textbf{95.00\%} & \textbf{98.33\%} & \textbf{96.67\%} & \textbf{90.00\%} & \textbf{0.10} & \textbf{0.22} & \textbf{0.25} & \textbf{0.90} & 55.32 & 59.71 & 67.06 & 73.77 \\
\hline
\multicolumn{13}{c}{\textbf{50 m Obstacle-Free Maps}}\\
\hline
\multirow{2}{*}{Method} & \multicolumn{4}{c|}{Success Rate $\uparrow$} & \multicolumn{4}{c|}{Solution Time/s $\downarrow$} & \multicolumn{4}{c}{Task Completion Time/s $\downarrow$}\\
& 10 & 15 & 20 & 25 & 10 & 15 & 20 & 25 & 10 & 15 & 20 & 25 \\
\hline
SCC   & 98.33\% & 98.33\% & 70.00\% & 10.00\% & 0.40 & 1.70 & 7.19 & 8.63  & 48.22 & 51.36 & 52.25 & 56.34 \\
PP    & 98.33\% & 96.67\% & 90.00\% & 73.33\% & 0.37 & 0.59 & 1.28 & 1.58  & 48.52 & 52.06 & 56.25 & 59.48 \\
FOTP  & 96.67\% & 88.33\% & 93.33\% & 85.00\% & 7.09 & 16.52 & 31.89 & 53.85 & 59.35 & 62.66 & 65.26 & 68.82 \\
SMART & \textbf{100.00\%} & \textbf{98.33\%} & \textbf{98.33\%} & \textbf{96.67\%} & \textbf{0.03} & \textbf{0.08} & \textbf{0.26} & \textbf{0.83} & 52.30 & 58.31 & 62.25 & 66.46 \\
\hline
\multicolumn{13}{c}{\textbf{100 m Indoor Maps}}\\
\hline
\multirow{2}{*}{Method} & \multicolumn{4}{c|}{Success Rate $\uparrow$} & \multicolumn{4}{c|}{Solution Time/s $\downarrow$} & \multicolumn{4}{c}{Task Completion Time/s $\downarrow$}\\
& 10 & 20 & 30 & 40 & 10 & 20 & 30 & 40 & 10 & 20 & 30 & 40 \\
\hline
SCC   & 93.33\% & 33.33\% & 00.00\% & 00.00\% & 4.43 & \textemdash & \textemdash & \textemdash & 133.02 & 115.01 & \textemdash & \textemdash \\
PP    & 100.00\% & 91.67\% & 61.67\% & 18.33\% & 2.11 & 1.34 & 0.59 & 0.50 & 135.41 & 142.00 & 131.73 & 127.35 \\
FOTP  & 30.00\% & 25.00\% & 25.00\% & 05.00\% & 7.42 & 35.98 & 88.29 & 154.61 & 93.65 & 111.30 & 110.83 & 111.62 \\
SMART & \textbf{100.00\%} & \textbf{98.33\%} & \textbf{80.00\%} & \textbf{60.00\%} & \textbf{1.19} & \textbf{3.64} & \textbf{5.49} & \textbf{7.86} & 171.78 & 187.37 & 171.77 & 181.57 \\
\hline
\end{tabular}
\end{adjustbox}
\end{table}


\begin{table}[htbp]
\centering
\scriptsize
\setlength{\tabcolsep}{4pt}
\renewcommand{\arraystretch}{1.1}
\caption{Comparison results on 100\,m Random Obstacle Maps and 100\,m Obstacle-Free Maps. Metrics include success rate ($\uparrow$), solution time/s ($\downarrow$), task completion time/s ($\downarrow$), and runtime of search/s.}
\label{tab:100m_combined_note}
\begin{adjustbox}{width=\linewidth}
\begin{tabular}{c|c|c|c|c}
\hline
\multicolumn{5}{c}{{\scriptsize \textbf{100 m Random Obstacle Maps}}} \\
\hline
\textbf{Agents} & \textbf{Success Rate $\uparrow$} & \textbf{Solution Time/s $\downarrow$} & \textbf{Task Completion Time/s $\downarrow$} & \textbf{Runtime/s} \\
\hline
25 & 93.33\% & 0.54 & 51.16 & 0.48 \\
30 & 96.67\% & 0.93 & 51.72 & 0.87 \\
35 & 96.67\% & 1.42 & 54.55 & 1.34 \\
40 & 100.00\% & 2.87 & 52.93 & 2.77 \\
50 & 95.00\% & 8.20 & 55.30 & 8.10 \\
60 & 65.00\% & 9.68 & 57.08 & 9.55 \\
70 & 20.00\% & 11.45 & 58.75 & 11.25 \\
80 & 11.67\% & 16.49 & 62.14 & 16.31 \\
90 & 1.67\%  & 25.00 & 61.00 & 24.72 \\
\hline
\multicolumn{5}{c}{{\scriptsize \textbf{100 m Obstacle-Free Maps}}} \\
\hline
\textbf{Agents} & \textbf{Success Rate $\uparrow$} & \textbf{Solution Time/s $\downarrow$} & \textbf{Task Completion Time/s $\downarrow$} & \textbf{Runtime/s} \\
\hline
25 & 98.33\% & 0.25 & 50.63 & 0.23 \\
30 & 98.33\% & 0.42 & 52.54 & 0.40 \\
35 & 100.00\% & 0.84 & 52.42 & 0.80 \\
40 & 100.00\% & 1.37 & 53.18 & 1.34 \\
50 & 98.33\% & 4.89 & 55.41 & 4.85 \\
60 & 66.67\% & 9.12 & 56.40 & 9.07 \\
70 & 18.33\% & 10.06 & 57.82 & 10.00 \\
80 & 6.67\%  & 9.62 & 58.75 & 9.52 \\
90 & 1.67\%  & 16.18 & 61.00 & 16.12 \\
\hline
\end{tabular}
\end{adjustbox}

\vspace{1ex}
\parbox{0.95\linewidth}{\scriptsize
\textit{Note.} Agents denotes the number of agents. Success Rate is reported in percent, all time metrics are in seconds. Random Obstacle Maps are sourced from \texttt{obstacle\_map100.csv}, Obstacle-Free Maps from \texttt{empty\_map100.csv}.
}
\end{table}

As shown in Table~\ref{tab:full_comparison_eng}, SMART achieves the highest
success rates across nearly all $50$\,m map scenarios. In random obstacle maps,
when the number of agents increases from 10 to 25, the success rates of PP and
FOTP degrade rapidly (down to 51.67\% for 25 vehicles), and SCC almost fails
completely (5\%). SMART, in contrast, maintains 90\% success even in these
dense cluttered environments. In obstacle-free settings, SMART again performs
best, reaching nearly perfect success rates up to 25 vehicles. For $100$\,m
indoor maps, where spatial constraints are more severe, SCC and FOTP fail once
the number of vehicles exceeds 20, while PP drops to 18.33\% at 40 vehicles.
SMART still sustains 80\% success at 30 vehicles and 60\% at 40 vehicles,
showing strong scalability.

Table~\ref{tab:100m_combined_note} further evaluates large-scale $100$\,m
scenarios with up to 90 vehicles. In both random obstacle and obstacle-free
maps, SMART shows clear advantages in maintaining higher success rates and
lower planning time and search time. Here, planning time refers to the overall
runtime required to generate feasible trajectories, while search time denotes the
front-end reasoning effort. For example, in random obstacle maps, success
remains above 90\% up to 50 vehicles, while competing methods collapse much
earlier. Even at 60--70 vehicles, SMART continues to achieve non-negligible
success (65\% and 20\%, respectively), whereas baselines almost completely fail.
A similar trend appears in obstacle-free maps, where SMART maintains more than
98\% success up to 50 vehicles and remains feasible even when the number of
vehicles approaches 90.

Computation time also highlights the advantages of SMART. In $50$\,m random
obstacle maps, the runtime stays below 1\,s even for 25 vehicles, which is at
least an order of magnitude faster than optimization-only baselines such as
FOTP. In $100$\,m random maps, SMART solves scenarios with 50 vehicles in only
8.20\,s, compared to tens or even hundreds of seconds required by baselines.
This efficiency gain results from the hierarchical design: the upper-level
search produces promising homotopies, while the distributed optimization
ensures feasibility and allows parallel computation.

Across more than 2000 test instances covering different map types, densities,
and scales, SMART provides the best balance of success rate, solution quality,
and runtime efficiency. These results confirm that the integration of large-step
priority search with distributed optimization improves both robustness in dense
environments and scalability to large fleets, outperforming state-of-the-art
methods in real-time multi-vehicle trajectory planning.

\subsection{Real-World Multi-Vehicle Trajectory Planning Experiments}

To further evaluate the practicality of the proposed algorithm in real-world
applications, we conduct experiments on a connected and automated vehicle
platform deployed in unstructured road environments. The experiments aim to
assess the responsiveness and trajectory planning performance of the algorithm
under complex traffic interactions, thereby validating its real-time feasibility
and effectiveness in physical scenarios.

\subsubsection{Hardware Platform}

\begin{figure}[htbp]
\centering
\includegraphics[width=1\textwidth]{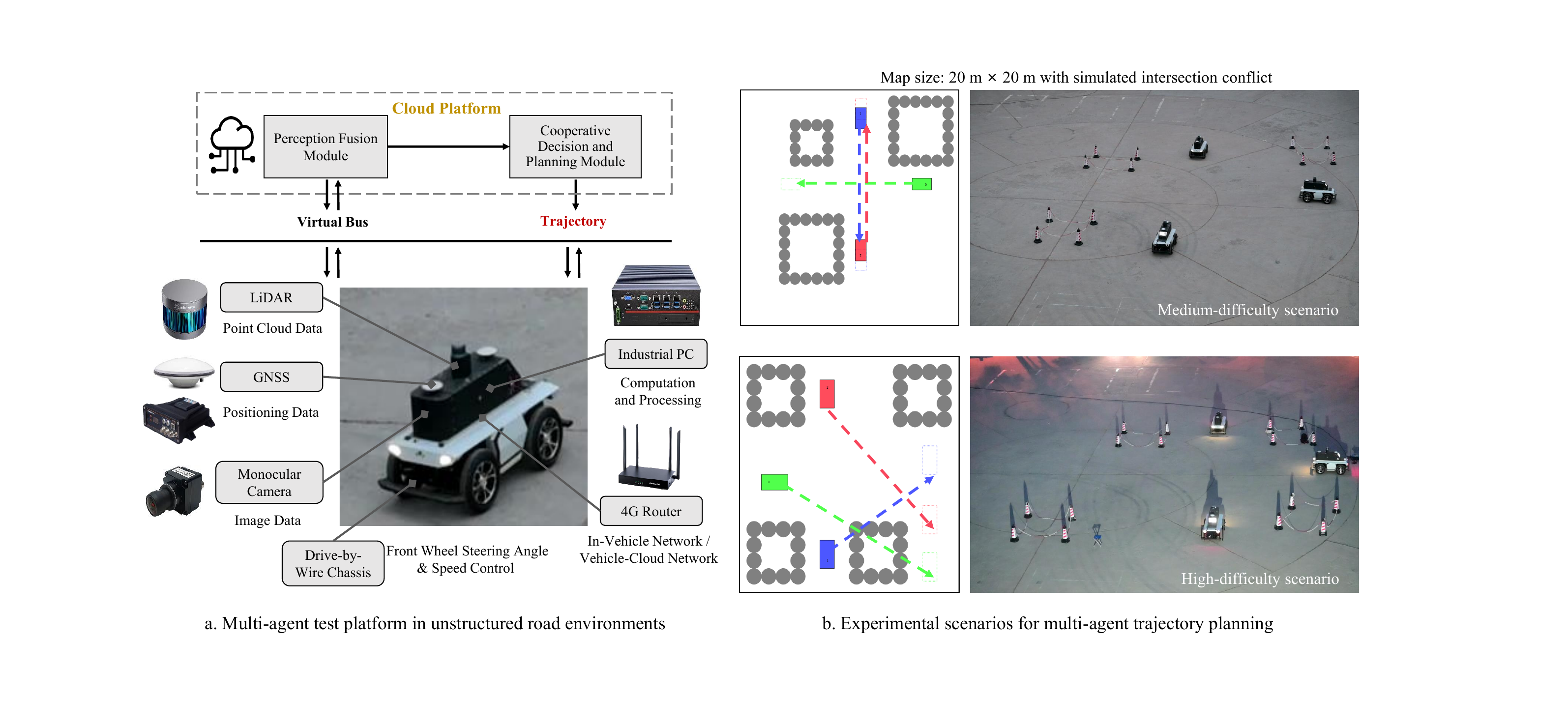}
\caption{Real-world multi-agent test platform and experimental scenarios.
(a) Experimental platform in unstructured road environments, showing onboard 
sensing, computation, and cloud-based modules. 
(b) Medium- and high-difficulty intersection scenarios used for trajectory planning experiments.}
\label{fig:fig6}
\end{figure}

A flexible real-world testbed is constructed using traffic cones to form obstacle maps 
of various sizes and layouts (Fig.~\ref{fig:fig6}a). All vehicles are connected and 
controllable, enabling the deployment of cooperative decision-making and planning 
algorithms in unstructured environments. Each vehicle is equipped with a 32-line LiDAR, 
a monocular camera, and an integrated navigation system. The LiDAR, mounted on top, 
provides 360-degree point cloud perception, while the camera delivers image data for 
scene understanding. An onboard industrial computer (Intel Core i7-8700 CPU and dual 
NVIDIA RTX 2080 Ti GPUs) supports stable execution of perception and planning modules 
in outdoor conditions. A 4G router enables communication with the cloud platform, and 
high-bandwidth data are transmitted through in-vehicle Ethernet. The chassis is 
Ackermann-steering with four-wheel drive, allowing precise maneuvering. Each vehicle 
runs perception and localization modules onboard and transmits fused perception 
results and state information (pose, heading, velocity) to the cloud in real time.

\subsubsection{Experimental Design}

The experiments emulate intersection-like conflict scenarios (Fig.~\ref{fig:fig6}b), where vehicles are placed symmetrically around the center with crossing start and goal states. For example, the upper and lower vehicles must exchange positions while the right-side vehicle crosses simultaneously, which would cause a collision at the center if all moved straight at constant speed. The task is thus to resolve these conflicts within a confined area by generating safe and coordinated trajectories. Two difficulty levels are considered: in the medium-difficulty case, wider corridors allow conflicts to be resolved through short detours or brief waiting; in the high-difficulty case, narrower corridors force tighter sequencing and more constrained maneuvers such as yielding or multi-stage avoidance. This design provides a systematic basis to evaluate SMART’s ability to adapt from moderately constrained to highly restricted environments.

\subsubsection{Experimental Results and Analysis}

\begin{figure}[htbp]
\centering
\includegraphics[width=1\textwidth]{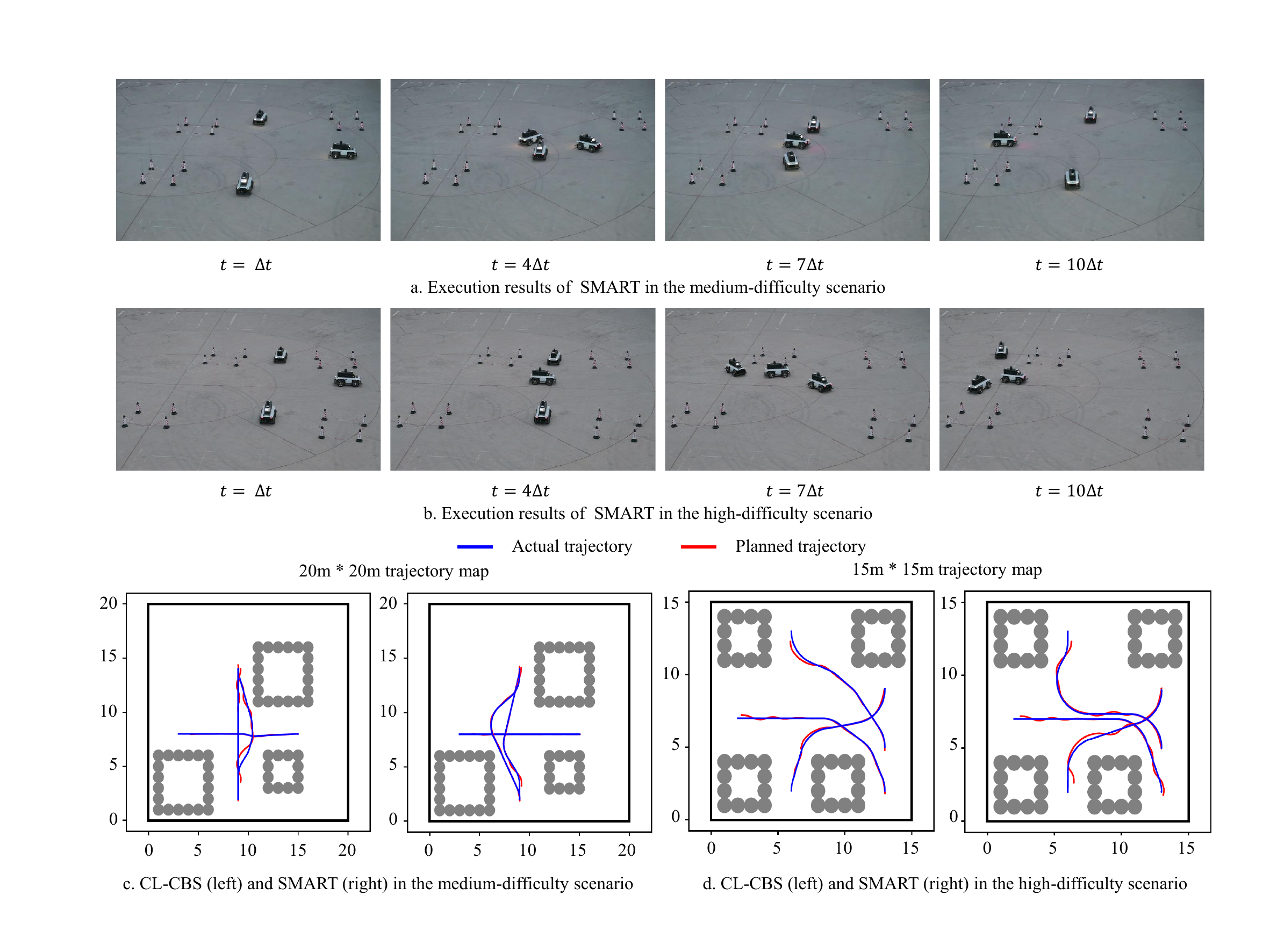}
\caption{Comparison of real-world results in medium- and high-difficulty scenarios.
(a) Execution of SMART in the $20\times20$\,m scenario at multiple time steps.
(b) Execution of SMART in the $15\times15$\,m scenario at the same time steps.
(c) Medium-difficulty comparison: CL\mbox{-}CBS (left) vs.\ SMART (right),
showing planned (blue) and executed (red) trajectories.
(d) High-difficulty comparison: CL\mbox{-}CBS (left) vs.\ SMART (right),
showing planned and executed trajectories.}
\label{fig:fig7}
\end{figure}

Figure~\ref{fig:fig7} presents a consolidated comparison of execution and planning
results in real-world tests under medium- and high-difficulty scenarios. The top row 
(a–b) shows time-lapse execution of \emph{SMART} at 
$t=\Delta t,\,4\Delta t,\,7\Delta t,\,10\Delta t$. 

In the medium-difficulty $20\times20$\,m case (Fig.~\ref{fig:fig7}a), corresponding to 
the scenario in Fig.~\ref{fig:fig6}b, the planned priority order $A \prec B \prec C$ is clearly reflected. 
Vehicle~A, located at the bottom of the scene, moves almost straight toward its goal 
without significant detours. Vehicle~B, starting from the top, adjusts its motion near the 
conflict point, temporarily detouring to the right to avoid A before steering back toward 
the center and continuing to its destination. Vehicle~C, starting on the right side, slows 
down and waits near the intersection point, yielding to both A and B before finally 
proceeding along its path. This behavior demonstrates how the algorithm enforces the 
priority sequence while enabling smooth conflict resolution among multiple vehicles.

In the high-difficulty $15\times15$\,m case (Fig.~\ref{fig:fig7}b), corresponding to 
the scenario in Fig.~\ref{fig:fig6}b, the environment imposes narrower corridors and stronger 
constraints. Vehicle~A, starting from the right edge, has the highest priority and quickly 
moves to its goal with a direct turning maneuver. Vehicle~B, at the top, must wait until 
A clears the conflict zone before advancing. It then performs a double-shift maneuver, 
temporarily adjusting its path twice to avoid A’s trajectory before moving forward. 
Vehicle~C, starting from the bottom, makes slight lateral adjustments, waiting for both 
A and B to pass, and then performs a similar double-shift maneuver to safely reach its 
destination. These cooperative patterns highlight that SMART not only preserves the 
designed priority ordering but also adapts to highly constrained layouts by enabling 
vehicles to yield, wait, and re-accelerate when necessary.

The bottom row (c–d) compares planned (blue) and executed (red)
trajectories between the strong baseline (CL\mbox{-}CBS, left) and
SMART (right).
In the medium-difficulty scenario (Fig.~\ref{fig:fig7}c),
CL\mbox{-}CBS requires $0.332$\,s for planning with a task completion
time of $15.5$\,s, while SMART computes in only $0.014$\,s ($\sim4\%$
of CL\mbox{-}CBS) with a task time of $18.1$\,s. In the high-difficulty
scenario (Fig.~\ref{fig:fig7}d), CL\mbox{-}CBS requires $10.075$\,s
(planning) and $17.4$\,s (task), whereas SMART computes in $0.623$\,s
($\sim6\%$ of CL\mbox{-}CBS) and completes in $20.3$\,s. Despite slightly 
longer execution times, SMART achieves an order-of-magnitude speedup in 
planning efficiency, demonstrating its real-time feasibility. Moreover, the 
executed trajectories remain close to the planned ones, confirming robustness 
against perception and actuation uncertainties in physical tests.

Overall, Fig.~\ref{fig:fig7} demonstrates that SMART consistently
achieves collision-free, cooperative solutions aligned with assigned 
priorities. Its ability to reduce computation time by more than $90\%$ 
compared to the baseline, while maintaining safe and interpretable 
execution, validates its effectiveness for real-world multi-vehicle coordination.

\section{Conclusion}
This paper introduced SMART, a hierarchical framework for scalable multi-vehicle trajectory planning in dense environments. By combining large-step priority-based search at the upper layer with distributed optimization at the lower layer, SMART bridges the gap between broad homotopy exploration and efficient local refinement, ensuring kinematic feasibility, collision avoidance, and computational efficiency. The upper layer leverages reinforcement learning-based priority estimation and tree search to explore diverse interaction modes, while the lower layer decomposes the coupled problem into parallelizable convex subproblems via corridor construction. This design ensures kinematic feasibility, collision avoidance, and computational efficiency.
Simulations highlighted SMART’s clear advantages. On $50$\,m $\times$ $50$\,m maps, it sustained over $90\%$ success for $25$ vehicles, while baselines such as SCC and FOTP dropped sharply. On $100$\,m $\times$ $100$\,m maps, SMART achieved above $95\%$ success up to $50$ vehicles and remained feasible at $90$ vehicles, where others nearly failed. Runtime analysis showed that SMART solved $50$-vehicle scenarios in $8.20$\,s, an order of magnitude faster than optimization-only baselines. 
Built on V2X communication, SMART also incorporates vehicle-infrastructure cooperation, where roadside sensing and coordination enhance multi-agent interaction. Real-vehicle field experiments further confirmed this design, reducing computation time by one to two orders of magnitude while preserving cooperative behaviors. Overall, SMART provides a robust and scalable solution for dense multi-vehicle trajectory planning empowered by V2X-enabled cooperation.
Future work will extend the framework to uncertain and mixed traffic environments, enhancing adaptability and robustness. A key direction is to develop more efficient methods for decision-making and coordination in large-scale scenarios with heterogeneous vehicles, leveraging V2X and infrastructure support to further strengthen system-level safety and efficiency.

\section*{CRediT authorship contribution statement}

Heye Huang: Conceptualization, Formal analysis, Writing – original draft, Visualization. 
Yibin Yang: Writing – review \& editing, Visualization, Validation. 
Wang Chen: Data curation, Investigation. 
Tiantian Chen: Formal analysis, Visualization, Investigation. 
Sikai Chen: Supervision, Conceptualization, Writing – review \& editing.

\section*{Declaration of competing interest}

The authors declare that they have no known competing financial interests or personal relationships that could have appeared to influence the work reported in this paper.

\section*{Acknowledgment}
\label{ack}

We gratefully acknowledge the THICV research group at the School of Vehicle and Mobility, Tsinghua University, for providing the experimental platform, computational resources, and support from all participants involved in the on-road vehicle experiments.

\section*{Data availability}

Data will be made available on request.

\bibliography{reference}

\begin{thebibliography}{39}
\expandafter\ifx\csname natexlab\endcsname\relax\def\natexlab#1{#1}\fi
\providecommand{\url}[1]{\texttt{#1}}
\providecommand{\href}[2]{#2}
\providecommand{\path}[1]{#1}
\providecommand{\DOIprefix}{doi:}
\providecommand{\ArXivprefix}{arXiv:}
\providecommand{\URLprefix}{URL: }
\providecommand{\Pubmedprefix}{pmid:}
\providecommand{\doi}[1]{\href{http://dx.doi.org/#1}{\path{#1}}}
\providecommand{\Pubmed}[1]{\href{pmid:#1}{\path{#1}}}
\providecommand{\bibinfo}[2]{#2}
\ifx\xfnm\relax \def\xfnm[#1]{\unskip,\space#1}\fi
\bibitem[{Allibhoy \& Cortés(2024)}]{allibhoy_control-barrier-function-based_2024}
\bibinfo{author}{Allibhoy, A.}, \& \bibinfo{author}{Cortés, J.} (\bibinfo{year}{2024}).
\newblock \bibinfo{title}{Control-{Barrier}-{Function}-{Based} {Design} of {Gradient} {Flows} for {Constrained} {Nonlinear} {Programming}}.
\newblock {\it \bibinfo{journal}{IEEE Transactions on Automatic Control}\/},  {\it \bibinfo{volume}{69}\/}, \bibinfo{pages}{3499--3514}.
\bibitem[{Benedikter et~al.(2019)Benedikter, Zavoli \& Colasurdo}]{benedikter2019convex}
\bibinfo{author}{Benedikter, B.}, \bibinfo{author}{Zavoli, A.}, \& \bibinfo{author}{Colasurdo, G.} (\bibinfo{year}{2019}).
\newblock \bibinfo{title}{A convex optimization approach for finite-thrust time-constrained cooperative rendezvous}.
\newblock {\it \bibinfo{journal}{arXiv preprint arXiv:1909.09443}\/}, .
\bibitem[{{\v{C}}{\'a}p et~al.(2015){\v{C}}{\'a}p, Nov{\'a}k, Kleiner \& Seleck{\`y}}]{vcap2015prioritized}
\bibinfo{author}{{\v{C}}{\'a}p, M.}, \bibinfo{author}{Nov{\'a}k, P.}, \bibinfo{author}{Kleiner, A.}, \& \bibinfo{author}{Seleck{\`y}, M.} (\bibinfo{year}{2015}).
\newblock \bibinfo{title}{Prioritized planning algorithms for trajectory coordination of multiple mobile robots}.
\newblock {\it \bibinfo{journal}{IEEE transactions on automation science and engineering}\/},  {\it \bibinfo{volume}{12}\/}, \bibinfo{pages}{835--849}.
\bibitem[{Chen et~al.(2015)Chen, Cutler \& How}]{chen2015decoupled}
\bibinfo{author}{Chen, Y.}, \bibinfo{author}{Cutler, M.}, \& \bibinfo{author}{How, J.~P.} (\bibinfo{year}{2015}).
\newblock \bibinfo{title}{Decoupled multiagent path planning via incremental sequential convex programming}.
\newblock In {\it \bibinfo{booktitle}{2015 IEEE International Conference on Robotics and Automation (ICRA)}\/} (pp. \bibinfo{pages}{5954--5961}).
\newblock \bibinfo{organization}{IEEE}.
\bibitem[{Dayan et~al.(2023)Dayan, Solovey, Pavone \& Halperin}]{dayan_near-optimal_2023}
\bibinfo{author}{Dayan, D.}, \bibinfo{author}{Solovey, K.}, \bibinfo{author}{Pavone, M.}, \& \bibinfo{author}{Halperin, D.} (\bibinfo{year}{2023}).
\newblock \bibinfo{title}{Near-{Optimal} {Multi}-{Robot} {Motion} {Planning} with {Finite} {Sampling}}.
\newblock {\it \bibinfo{journal}{IEEE Transactions on Robotics}\/},  {\it \bibinfo{volume}{39}\/}, \bibinfo{pages}{3422--3436}.
\bibitem[{Huang et~al.(2025)Huang, Liu, Zhang, Zhao, Li \& Wang}]{huang_lead_2025}
\bibinfo{author}{Huang, H.}, \bibinfo{author}{Liu, J.}, \bibinfo{author}{Zhang, B.}, \bibinfo{author}{Zhao, S.}, \bibinfo{author}{Li, B.}, \& \bibinfo{author}{Wang, J.} (\bibinfo{year}{2025}).
\newblock \bibinfo{title}{{LEAD}: {Learning}-{Enhanced} {Adaptive} {Decision}-{Making} for {Autonomous} {Driving} in {Dynamic} {Environments}}.
\newblock {\it \bibinfo{journal}{IEEE Transactions on Intelligent Transportation Systems}\/},  {\it \bibinfo{volume}{26}\/}, \bibinfo{pages}{6142--6156}.
\bibitem[{Huang et~al.(2024)Huang, Liu, Liu, Yang, Wang, Abbink \& Zgonnikov}]{huang_general_2024}
\bibinfo{author}{Huang, H.}, \bibinfo{author}{Liu, Y.}, \bibinfo{author}{Liu, J.}, \bibinfo{author}{Yang, Q.}, \bibinfo{author}{Wang, J.}, \bibinfo{author}{Abbink, D.}, \& \bibinfo{author}{Zgonnikov, A.} (\bibinfo{year}{2024}).
\newblock \bibinfo{title}{General {Optimal} {Trajectory} {Planning}: {Enabling} {Autonomous} {Vehicles} with the {Principle} of {Least} {Action}}.
\newblock {\it \bibinfo{journal}{Engineering}\/},  {\it \bibinfo{volume}{33}\/}, \bibinfo{pages}{63--76}.
\bibitem[{Lee et~al.(2025)Lee, Park \& Kim}]{lee_dmvc-tracker_2025}
\bibinfo{author}{Lee, Y.}, \bibinfo{author}{Park, J.}, \& \bibinfo{author}{Kim, H.~J.} (\bibinfo{year}{2025}).
\newblock \bibinfo{title}{{DMVC}-{Tracker}: {Distributed} {Multi}-{Agent} {Trajectory} {Planning} for {Target} {Tracking} {Using} {Dynamic} {Buffered} {Voronoi} and {Inter}-{Visibility} {Cells}}.
\newblock {\it \bibinfo{journal}{IEEE Robotics and Automation Letters}\/},  {\it \bibinfo{volume}{10}\/}, \bibinfo{pages}{4842--4849}.
\bibitem[{Li et~al.(2021)Li, Ouyang, Zhang, Acarman, Kong \& Shao}]{li_optimal_2021}
\bibinfo{author}{Li, B.}, \bibinfo{author}{Ouyang, Y.}, \bibinfo{author}{Zhang, Y.}, \bibinfo{author}{Acarman, T.}, \bibinfo{author}{Kong, Q.}, \& \bibinfo{author}{Shao, Z.} (\bibinfo{year}{2021}).
\newblock \bibinfo{title}{Optimal {Cooperative} {Maneuver} {Planning} for {Multiple} {Nonholonomic} {Robots} in a {Tiny} {Environment} via {Adaptive}-{Scaling} {Constrained} {Optimization}}.
\newblock {\it \bibinfo{journal}{IEEE Robotics and Automation Letters}\/},  {\it \bibinfo{volume}{6}\/}, \bibinfo{pages}{1511--1518}.
\bibitem[{Lin et~al.(2025)Lin, Song, Zhu \& Zhu}]{lin_multi-agent_2025}
\bibinfo{author}{Lin, W.}, \bibinfo{author}{Song, W.}, \bibinfo{author}{Zhu, Q.}, \& \bibinfo{author}{Zhu, S.} (\bibinfo{year}{2025}).
\newblock \bibinfo{title}{Multi-{Agent} {Path} {Finding} {With} {Heterogeneous} {Geometric} and {Kinematic} {Constraints} in {Continuous} {Space}}.
\newblock {\it \bibinfo{journal}{IEEE Robotics and Automation Letters}\/},  {\it \bibinfo{volume}{10}\/}, \bibinfo{pages}{492--499}.
\bibitem[{Liu et~al.(2024)Liu, Huang, Xu, Xu \& Wang}]{liu2024safer}
\bibinfo{author}{Liu, Y.}, \bibinfo{author}{Huang, H.}, \bibinfo{author}{Xu, Q.}, \bibinfo{author}{Xu, S.}, \& \bibinfo{author}{Wang, J.} (\bibinfo{year}{2024}).
\newblock \bibinfo{title}{Safer conflict-based search: Risk-constrained optimal pathfinding for multiple connected and automated vehicles}.
\newblock {\it \bibinfo{journal}{IEEE Transactions on Automation Science and Engineering}\/}, .
\bibitem[{Lukyanenko \& Soudbakhsh(2023)}]{lukyanenko_probabilistic_2023}
\bibinfo{author}{Lukyanenko, A.}, \& \bibinfo{author}{Soudbakhsh, D.} (\bibinfo{year}{2023}).
\newblock \bibinfo{title}{Probabilistic motion planning for non-{Euclidean} and multi-vehicle problems}.
\newblock {\it \bibinfo{journal}{Robotics and Autonomous Systems}\/},  {\it \bibinfo{volume}{168}\/}, \bibinfo{pages}{104487}.
\bibitem[{Ma et~al.(2017)Ma, Li, Kumar \& Koenig}]{ma_lifelong_2017}
\bibinfo{author}{Ma, H.}, \bibinfo{author}{Li, J.}, \bibinfo{author}{Kumar, T. K.~S.}, \& \bibinfo{author}{Koenig, S.} (\bibinfo{year}{2017}).
\newblock \bibinfo{title}{Lifelong {Multi}-{Agent} {Path} {Finding} for {Online} {Pickup} and {Delivery} {Tasks}}.
\newblock \bibinfo{note}{ArXiv:1705.10868 [cs]}.
\bibitem[{Marcucci(2024)}]{marcucci2024graphs}
\bibinfo{author}{Marcucci, T.} (\bibinfo{year}{2024}).
\newblock {\it \bibinfo{title}{Graphs of convex sets with applications to optimal control and motion planning}\/}.
\newblock Ph.D. thesis Massachusetts Institute of Technology.
\bibitem[{Mellinger \& Kumar(2011)}]{mellinger_minimum_2011}
\bibinfo{author}{Mellinger, D.}, \& \bibinfo{author}{Kumar, V.} (\bibinfo{year}{2011}).
\newblock \bibinfo{title}{Minimum snap trajectory generation and control for quadrotors}.
\newblock In {\it \bibinfo{booktitle}{2011 {IEEE} {International} {Conference} on {Robotics} and {Automation}}\/} (pp. \bibinfo{pages}{2520--2525}).
\newblock \bibinfo{note}{ISSN: 1050-4729}.
\bibitem[{Meng et~al.(2025)Meng, Zhang, Zhou, Guo \& Hu}]{meng2025advances}
\bibinfo{author}{Meng, W.}, \bibinfo{author}{Zhang, X.}, \bibinfo{author}{Zhou, L.}, \bibinfo{author}{Guo, H.}, \& \bibinfo{author}{Hu, X.} (\bibinfo{year}{2025}).
\newblock \bibinfo{title}{Advances in uav path planning: A comprehensive review of methods, challenges, and future directions.}
\newblock {\it \bibinfo{journal}{Drones (2504-446X)}\/},  {\it \bibinfo{volume}{9}\/}.
\bibitem[{Nascimento et~al.(2016)Nascimento, Conceição \& Moreira}]{nascimento_multi-robot_2016}
\bibinfo{author}{Nascimento, T.~P.}, \bibinfo{author}{Conceição, A. G.~S.}, \& \bibinfo{author}{Moreira, A.~P.} (\bibinfo{year}{2016}).
\newblock \bibinfo{title}{Multi-{Robot} nonlinear model predictive formation control: the obstacle avoidance problem}.
\newblock {\it \bibinfo{journal}{Robotica}\/},  {\it \bibinfo{volume}{34}\/}, \bibinfo{pages}{549--567}.
\bibitem[{Nikou et~al.(2020)Nikou, Heshmati-alamdari \& Dimarogonas}]{nikou_scalable_2020}
\bibinfo{author}{Nikou, A.}, \bibinfo{author}{Heshmati-alamdari, S.}, \& \bibinfo{author}{Dimarogonas, D.~V.} (\bibinfo{year}{2020}).
\newblock \bibinfo{title}{Scalable time-constrained planning of multi-robot systems}.
\newblock {\it \bibinfo{journal}{Autonomous Robots}\/},  {\it \bibinfo{volume}{44}\/}, \bibinfo{pages}{1451--1467}.
\bibitem[{Ouyang et~al.(2022{\natexlab{a}})Ouyang, Li, Zhang, Acarman, Guo \& Zhang}]{ouyang_fast_2022}
\bibinfo{author}{Ouyang, Y.}, \bibinfo{author}{Li, B.}, \bibinfo{author}{Zhang, Y.}, \bibinfo{author}{Acarman, T.}, \bibinfo{author}{Guo, Y.}, \& \bibinfo{author}{Zhang, T.} (\bibinfo{year}{2022}{\natexlab{a}}).
\newblock \bibinfo{title}{Fast and {Optimal} {Trajectory} {Planning} for {Multiple} {Vehicles} in a {Nonconvex} and {Cluttered} {Environment}: {Benchmarks}, {Methodology}, and {Experiments}}.
\newblock In {\it \bibinfo{booktitle}{2022 {International} {Conference} on {Robotics} and {Automation} ({ICRA})}\/} (pp. \bibinfo{pages}{10746--10752}).
\bibitem[{Ouyang et~al.(2022{\natexlab{b}})Ouyang, Li, Zhang, Acarman, Guo \& Zhang}]{ouyang2022fast}
\bibinfo{author}{Ouyang, Y.}, \bibinfo{author}{Li, B.}, \bibinfo{author}{Zhang, Y.}, \bibinfo{author}{Acarman, T.}, \bibinfo{author}{Guo, Y.}, \& \bibinfo{author}{Zhang, T.} (\bibinfo{year}{2022}{\natexlab{b}}).
\newblock \bibinfo{title}{Fast and optimal trajectory planning for multiple vehicles in a nonconvex and cluttered environment: Benchmarks, methodology, and experiments}.
\newblock In {\it \bibinfo{booktitle}{2022 International Conference on Robotics and Automation (ICRA)}\/} (pp. \bibinfo{pages}{10746--10752}).
\newblock \bibinfo{organization}{IEEE}.
\bibitem[{Park et~al.(2022)Park, Kim, Kim, Oh \& Kim}]{park_online_2022}
\bibinfo{author}{Park, J.}, \bibinfo{author}{Kim, D.}, \bibinfo{author}{Kim, G.~C.}, \bibinfo{author}{Oh, D.}, \& \bibinfo{author}{Kim, H.~J.} (\bibinfo{year}{2022}).
\newblock \bibinfo{title}{Online {Distributed} {Trajectory} {Planning} for {Quadrotor} {Swarm} {With} {Feasibility} {Guarantee} {Using} {Linear} {Safe} {Corridor}}.
\newblock {\it \bibinfo{journal}{IEEE Robotics and Automation Letters}\/},  {\it \bibinfo{volume}{7}\/}, \bibinfo{pages}{4869--4876}.
\bibitem[{Park et~al.(2020{\natexlab{a}})Park, Kim, Jang \& Kim}]{park_efficient_2020}
\bibinfo{author}{Park, J.}, \bibinfo{author}{Kim, J.}, \bibinfo{author}{Jang, I.}, \& \bibinfo{author}{Kim, H.~J.} (\bibinfo{year}{2020}{\natexlab{a}}).
\newblock \bibinfo{title}{Efficient {Multi}-{Agent} {Trajectory} {Planning} with {Feasibility} {Guarantee} using {Relative} {Bernstein} {Polynomial}}.
\newblock In {\it \bibinfo{booktitle}{2020 {IEEE} {International} {Conference} on {Robotics} and {Automation} ({ICRA})}\/} (pp. \bibinfo{pages}{434--440}).
\newblock \bibinfo{note}{ISSN: 2577-087X}.
\bibitem[{Park et~al.(2020{\natexlab{b}})Park, Kim, Jang \& Kim}]{park_efficient_2020-1}
\bibinfo{author}{Park, J.}, \bibinfo{author}{Kim, J.}, \bibinfo{author}{Jang, I.}, \& \bibinfo{author}{Kim, H.~J.} (\bibinfo{year}{2020}{\natexlab{b}}).
\newblock \bibinfo{title}{Efficient {Multi}-{Agent} {Trajectory} {Planning} with {Feasibility} {Guarantee} using {Relative} {Bernstein} {Polynomial}}.
\newblock In {\it \bibinfo{booktitle}{2020 {IEEE} {International} {Conference} on {Robotics} and {Automation} ({ICRA})}\/} (pp. \bibinfo{pages}{434--440}).
\newblock \bibinfo{note}{ISSN: 2577-087X}.
\bibitem[{Reis et~al.(2021)Reis, Aguiar \& Tabuada}]{reis_control_2021}
\bibinfo{author}{Reis, M.~F.}, \bibinfo{author}{Aguiar, A.~P.}, \& \bibinfo{author}{Tabuada, P.} (\bibinfo{year}{2021}).
\newblock \bibinfo{title}{Control {Barrier} {Function}-{Based} {Quadratic} {Programs} {Introduce} {Undesirable} {Asymptotically} {Stable} {Equilibria}}.
\newblock {\it \bibinfo{journal}{IEEE Control Systems Letters}\/},  {\it \bibinfo{volume}{5}\/}, \bibinfo{pages}{731--736}.
\bibitem[{Rey et~al.(2018)Rey, Pan, Hauswirth \& Lygeros}]{rey_fully_2018}
\bibinfo{author}{Rey, F.}, \bibinfo{author}{Pan, Z.}, \bibinfo{author}{Hauswirth, A.}, \& \bibinfo{author}{Lygeros, J.} (\bibinfo{year}{2018}).
\newblock \bibinfo{title}{Fully {Decentralized} {ADMM} for {Coordination} and {Collision} {Avoidance}}.
\newblock In {\it \bibinfo{booktitle}{2018 {European} {Control} {Conference} ({ECC})}\/} (pp. \bibinfo{pages}{825--830}).
\bibitem[{Schouwenaars et~al.(2001{\natexlab{a}})Schouwenaars, De~Moor, Feron \& How}]{schouwenaars_mixed_2001}
\bibinfo{author}{Schouwenaars, T.}, \bibinfo{author}{De~Moor, B.}, \bibinfo{author}{Feron, E.}, \& \bibinfo{author}{How, J.} (\bibinfo{year}{2001}{\natexlab{a}}).
\newblock \bibinfo{title}{Mixed integer programming for multi-vehicle path planning}.
\newblock In {\it \bibinfo{booktitle}{2001 {European} {Control} {Conference} ({ECC})}\/} (pp. \bibinfo{pages}{2603--2608}).
\bibitem[{Schouwenaars et~al.(2001{\natexlab{b}})Schouwenaars, De~Moor, Feron \& How}]{schouwenaars_mixed_2001-1}
\bibinfo{author}{Schouwenaars, T.}, \bibinfo{author}{De~Moor, B.}, \bibinfo{author}{Feron, E.}, \& \bibinfo{author}{How, J.} (\bibinfo{year}{2001}{\natexlab{b}}).
\newblock \bibinfo{title}{Mixed integer programming for multi-vehicle path planning}.
\newblock In {\it \bibinfo{booktitle}{2001 {European} {Control} {Conference} ({ECC})}\/} (pp. \bibinfo{pages}{2603--2608}).
\bibitem[{Shi et~al.(2022)Shi, Hönig, Shi, Yue \& Chung}]{shi_neural-swarm2_2022}
\bibinfo{author}{Shi, G.}, \bibinfo{author}{Hönig, W.}, \bibinfo{author}{Shi, X.}, \bibinfo{author}{Yue, Y.}, \& \bibinfo{author}{Chung, S.-J.} (\bibinfo{year}{2022}).
\newblock \bibinfo{title}{Neural-{Swarm2}: {Planning} and {Control} of {Heterogeneous} {Multirotor} {Swarms} {Using} {Learned} {Interactions}}.
\newblock {\it \bibinfo{journal}{IEEE Transactions on Robotics}\/},  {\it \bibinfo{volume}{38}\/}, \bibinfo{pages}{1063--1079}.
\bibitem[{Shome et~al.(2020)Shome, Solovey, Dobson, Halperin \& Bekris}]{shome_drrt_2020}
\bibinfo{author}{Shome, R.}, \bibinfo{author}{Solovey, K.}, \bibinfo{author}{Dobson, A.}, \bibinfo{author}{Halperin, D.}, \& \bibinfo{author}{Bekris, K.~E.} (\bibinfo{year}{2020}).
\newblock \bibinfo{title}{{dRRT}*: {Scalable} and informed asymptotically-optimal multi-robot motion planning}.
\newblock {\it \bibinfo{journal}{Autonomous Robots}\/},  {\it \bibinfo{volume}{44}\/}, \bibinfo{pages}{443--467}.
\bibitem[{Theurkauf et~al.(2024)Theurkauf, Kottinger, Ahmed \& Lahijanian}]{theurkauf_chance-constrained_2024}
\bibinfo{author}{Theurkauf, A.}, \bibinfo{author}{Kottinger, J.}, \bibinfo{author}{Ahmed, N.}, \& \bibinfo{author}{Lahijanian, M.} (\bibinfo{year}{2024}).
\newblock \bibinfo{title}{Chance-{Constrained} {Multi}-{Robot} {Motion} {Planning} {Under} {Gaussian} {Uncertainties}}.
\newblock {\it \bibinfo{journal}{IEEE Robotics and Automation Letters}\/},  {\it \bibinfo{volume}{9}\/}, \bibinfo{pages}{835--842}.
\bibitem[{Tordesillas \& How(2022)}]{tordesillas_mader_2022}
\bibinfo{author}{Tordesillas, J.}, \& \bibinfo{author}{How, J.~P.} (\bibinfo{year}{2022}).
\newblock \bibinfo{title}{{MADER}: {Trajectory} {Planner} in {Multiagent} and {Dynamic} {Environments}}.
\newblock {\it \bibinfo{journal}{IEEE Transactions on Robotics}\/},  {\it \bibinfo{volume}{38}\/}, \bibinfo{pages}{463--476}.
\bibitem[{Wang et~al.(2016)Wang, Ames \& Egerstedt}]{wang_safety_2016}
\bibinfo{author}{Wang, L.}, \bibinfo{author}{Ames, A.}, \& \bibinfo{author}{Egerstedt, M.} (\bibinfo{year}{2016}).
\newblock \bibinfo{title}{Safety barrier certificates for heterogeneous multi-robot systems}.
\newblock In {\it \bibinfo{booktitle}{2016 {American} {Control} {Conference} ({ACC})}\/} (pp. \bibinfo{pages}{5213--5218}).
\newblock \bibinfo{note}{ISSN: 2378-5861}.
\bibitem[{Wen et~al.(2022)Wen, Liu \& Li}]{wen_cl-mapf_2022}
\bibinfo{author}{Wen, L.}, \bibinfo{author}{Liu, Y.}, \& \bibinfo{author}{Li, H.} (\bibinfo{year}{2022}).
\newblock \bibinfo{title}{{CL}-{MAPF}: {Multi}-{Agent} {Path} {Finding} for {Car}-{Like} robots with kinematic and spatiotemporal constraints}.
\newblock {\it \bibinfo{journal}{Robotics and Autonomous Systems}\/},  {\it \bibinfo{volume}{150}\/}, \bibinfo{pages}{103997}.
\bibitem[{Yang et~al.(2023)Yang, Ai, Teng, Gao, Cui, Tian \& Chen}]{yang2023decoupled}
\bibinfo{author}{Yang, Q.}, \bibinfo{author}{Ai, Y.}, \bibinfo{author}{Teng, S.}, \bibinfo{author}{Gao, Y.}, \bibinfo{author}{Cui, C.}, \bibinfo{author}{Tian, B.}, \& \bibinfo{author}{Chen, L.} (\bibinfo{year}{2023}).
\newblock \bibinfo{title}{Decoupled real-time trajectory planning for multiple autonomous mining trucks in unloading areas}.
\newblock {\it \bibinfo{journal}{IEEE Transactions on Intelligent Vehicles}\/},  {\it \bibinfo{volume}{8}\/}, \bibinfo{pages}{4319--4330}.
\bibitem[{Yang et~al.(2024{\natexlab{a}})Yang, Fan, He, Wang, Huang \& Sartoretti}]{yang2024attention}
\bibinfo{author}{Yang, Y.}, \bibinfo{author}{Fan, M.}, \bibinfo{author}{He, C.}, \bibinfo{author}{Wang, J.}, \bibinfo{author}{Huang, H.}, \& \bibinfo{author}{Sartoretti, G.} (\bibinfo{year}{2024}{\natexlab{a}}).
\newblock \bibinfo{title}{Attention-based priority learning for limited time multi-agent path finding}.
\newblock In {\it \bibinfo{booktitle}{Proceedings of the 23rd International Conference on Autonomous Agents and Multiagent Systems}\/} (pp. \bibinfo{pages}{1993--2001}).
\bibitem[{Yang et~al.(2024{\natexlab{b}})Yang, Xu, Yan, Jiang, Wang \& Huang}]{yang2024csdo}
\bibinfo{author}{Yang, Y.}, \bibinfo{author}{Xu, S.}, \bibinfo{author}{Yan, X.}, \bibinfo{author}{Jiang, J.}, \bibinfo{author}{Wang, J.}, \& \bibinfo{author}{Huang, H.} (\bibinfo{year}{2024}{\natexlab{b}}).
\newblock \bibinfo{title}{Csdo: Enhancing efficiency and success in large-scale multi-vehicle trajectory planning}.
\newblock {\it \bibinfo{journal}{IEEE Robotics and Automation Letters}\/}, .
\bibitem[{Zhang et~al.(2024{\natexlab{a}})Zhang, Li, Zhang, Wang, Kwan~Ng \& Xu}]{zhang_multi-uncertainty_2024}
\bibinfo{author}{Zhang, S.}, \bibinfo{author}{Li, H.}, \bibinfo{author}{Zhang, S.}, \bibinfo{author}{Wang, S.}, \bibinfo{author}{Kwan~Ng, D.~W.}, \& \bibinfo{author}{Xu, C.} (\bibinfo{year}{2024}{\natexlab{a}}).
\newblock \bibinfo{title}{Multi-{Uncertainty} {Aware} {Autonomous} {Cooperative} {Planning}}.
\newblock In {\it \bibinfo{booktitle}{2024 {IEEE}/{RSJ} {International} {Conference} on {Intelligent} {Robots} and {Systems} ({IROS})}\/} (pp. \bibinfo{pages}{1018--1025}).
\newblock \bibinfo{note}{ISSN: 2153-0866}.
\bibitem[{Zhang et~al.(2024{\natexlab{b}})Zhang, Xiong, Wang, Teng \& Chen}]{zhang_d-pbs_2024}
\bibinfo{author}{Zhang, X.}, \bibinfo{author}{Xiong, G.}, \bibinfo{author}{Wang, Y.}, \bibinfo{author}{Teng, S.}, \& \bibinfo{author}{Chen, L.} (\bibinfo{year}{2024}{\natexlab{b}}).
\newblock \bibinfo{title}{D-{PBS}: {Dueling} {Priority}-{Based} {Search} for {Multiple} {Nonholonomic} {Robots} {Motion} {Planning} in {Congested} {Environments}}.
\newblock {\it \bibinfo{journal}{IEEE Robotics and Automation Letters}\/},  {\it \bibinfo{volume}{9}\/}, \bibinfo{pages}{6288--6295}.
\bibitem[{Şenbaşlar et~al.(2019)Şenbaşlar, Hönig \& Ayanian}]{senbaslar_robust_2019}
\bibinfo{author}{Şenbaşlar, B.}, \bibinfo{author}{Hönig, W.}, \& \bibinfo{author}{Ayanian, N.} (\bibinfo{year}{2019}).
\newblock \bibinfo{title}{Robust {Trajectory} {Execution} for {Multi}-robot {Teams} {Using} {Distributed} {Real}-time {Replanning}}.
\newblock In \bibinfo{editor}{N.~Correll}, \bibinfo{editor}{M.~Schwager}, \& \bibinfo{editor}{M.~Otte} (Eds.), {\it \bibinfo{booktitle}{Distributed {Autonomous} {Robotic} {Systems}}\/} (pp. \bibinfo{pages}{167--181}).
\newblock \bibinfo{address}{Cham}: \bibinfo{publisher}{Springer International Publishing}.

\end{thebibliography}

\end{document}